\documentclass[conference]{IEEEtran}

\usepackage{moreverb,url}
\usepackage[colorlinks,bookmarksopen,bookmarksnumbered,citecolor=red,urlcolor=red]{hyperref}
\usepackage{xcolor,soul,framed} 
\usepackage{multirow}
\colorlet{shadecolor}{yellow}
\usepackage{array}
\usepackage{mdwmath}
\usepackage{mdwtab}
\usepackage{eqparbox}
\usepackage{url}
\usepackage{enumerate}
\usepackage{amssymb}
\usepackage{rotating}
\usepackage{placeins}
\usepackage[caption=false,font=footnotesize]{subfig} 
\usepackage{graphicx}
\usepackage{adjustbox}
\usepackage{makecell}
\usepackage{pifont}
\usepackage{amsmath}
\usepackage{booktabs}
\usepackage{rotating}

\begin{document}

\title{Online Clustering of Seafloor Imagery for Interpretation during Long-Term AUV Operations\\
}

\author{
\IEEEauthorblockN{Cailei Liang}
\IEEEauthorblockA{
  \textit{University of Southampton}\\
  Southampton, UK \\
  c.liang@soton.ac.uk
}
\and
\IEEEauthorblockN{Adrian Bodenmann}
\IEEEauthorblockA{
  \textit{University of Southampton}\\
  Southampton, UK \\
  adrian.bodenmann@soton.ac.uk
}
\and
\IEEEauthorblockN{Sam Fenton}
\IEEEauthorblockA{
  \textit{University of Southampton}\\
  Southampton, UK \\
  s.n.fenton@soton.ac.uk
}
\and
\IEEEauthorblockN{Blair Thornton}
\IEEEauthorblockA{
  \textit{University of Southampton, UK}\\
  \textit{IIS, The University of Tokyo, Japan}\\
  b.thornton@soton.ac.uk
}


}  





\maketitle

\begin{abstract}
As long-endurance and seafloor-resident AUVs become more capable, there is an increasing need for extended, real-time interpretation of seafloor imagery to enable adaptive missions and optimise communication efficiency. Although offline image analysis methods are well established, they rely on access to complete datasets and human-labelled examples to manage the strong influence of environmental and operational conditions on seafloor image appearance—requirements that cannot be met in real-time settings. To address this, we introduce an online clustering framework (OCF) capable of interpreting seafloor imagery without supervision, that is designed to operate in real-time on continuous data streams in a scalable, adaptive, and self-consistent manner. The method enables the efficient review and consolidation of common patterns across the entire data history in constant time by identifying and maintaining a set of representative samples that capture the evolving feature distribution, supporting dynamic cluster merging and splitting without reprocessing the full image history. We evaluate the framework on three diverse seafloor image datasets, analysing the impact of different representative sampling strategies on both clustering accuracy and computational cost. The OCF achieves the highest average F1 score of 0.68 across the three datasets among all comparative online clustering approaches, with a standard deviation of 3\% across three distinct survey trajectories, demonstrating its superior clustering capability and robustness to trajectory variation. In addition, it maintains consistently lower and bounded computational time as the data volume increases. Compared to offline clustering methods, it strikes a favourable balance between accuracy and efficiency. These properties are beneficial for generating survey data summaries and supporting informative path planning in long-term, persistent autonomous marine exploration.

\end{abstract}


%

\IEEEpeerreviewmaketitle

\section{Introduction}

\begin{figure*}
    \centering
    \includegraphics[width=0.95\linewidth]{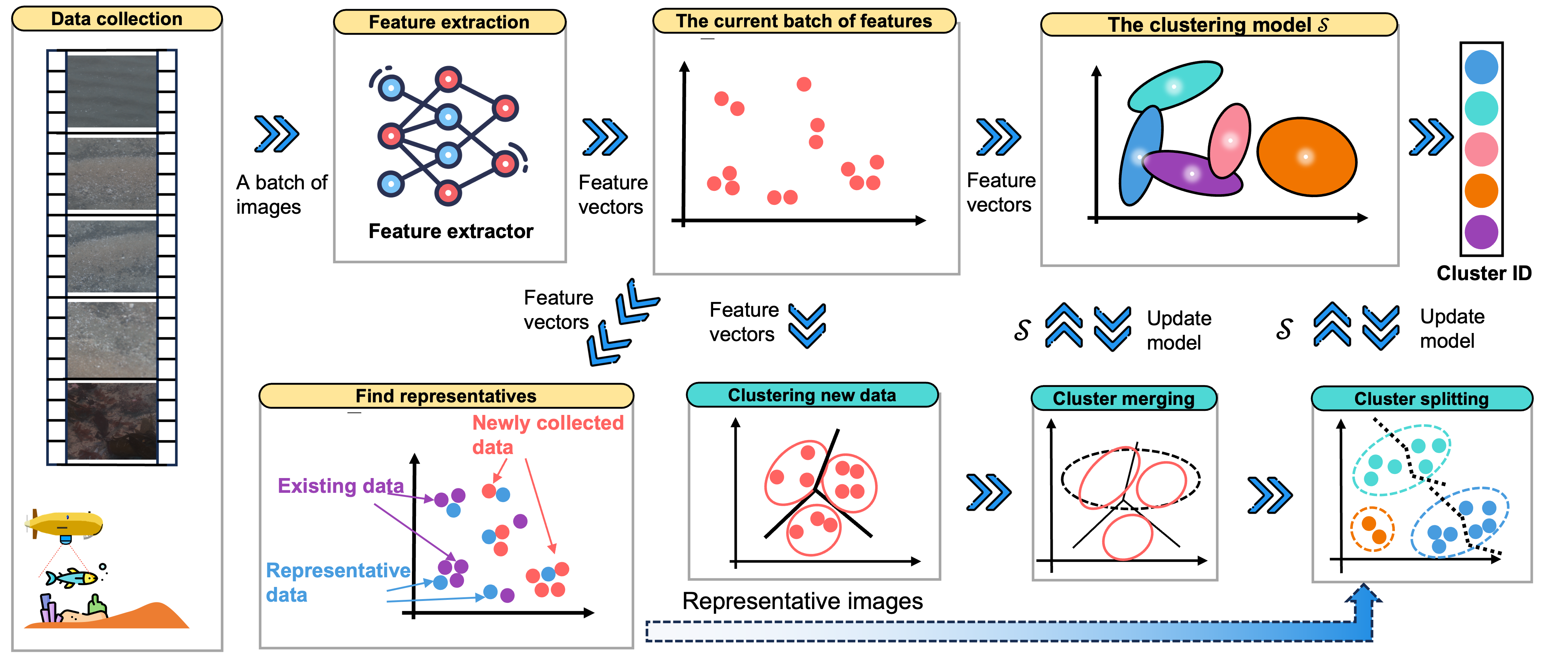}
    \caption{The OCF incrementally assigns images to a history-maintained cluster set $\mathcal{S}$. Each batch’s features are clustered and merged with existing ones, and they also contribute to updating a representative subset summarising the full data distribution. These representatives drive cluster splitting, and the updated $\mathcal{S}$ is then used to label the images.
}
    \label{fig:method_overview}
\end{figure*}

\IEEEPARstart{A}{utonomous} Underwater Vehicles (AUVs) are widely used to collect high-resolution images of the seafloor, enabling a wide range of applications including scientific exploration, infrastructure inspection and marine conservation~\cite{williams,nauert2023inspection}. Missions typically involve AUVs following pre-defined trajectories, with the acquired data processed offline to reconstruct large-scale spatial patterns in the observed environment~\cite{singh, steinberg2015hierarchical, yamada2021geoclr,bodenmann2017generation}. The strong attenuation of light in water requires seafloor images to be taken from close range (1-10\,m), limiting each image extent to a few metres. This is significantly less than the spatial scales over which seafloor habitats and substrate types extend (hundreds of metres~\cite{zelada}), meaning that new patterns are inevitably encountered and previously seen patterns can be revisited as missions progress. Real-time analysis can enhance survey efficiency by adapting AUV trajectories in response to patterns detected in observations, enabling more targeted and effective data collection~\cite{hwang2019auv}, or by communicating observation summaries to support timely operator decision making~\cite{murphy2013}. However, significant challenges remain. Zero-shot frameworks, where classifiers pre-trained on existing databases are applied to new data, have been developed for terrestrial applications, but do not exist for the diverse, application-specific label dictionaries used in subsea survey~\cite{oquab2023dinov2,
catami}. Since the appearance of seafloor images is highly sensitive to variations in lighting, vehicle altitude and turbidity, features extracted from similar scenes are prone to shift across class boundaries and degrade model performance~\cite{Langenkamper}. Although self-supervised and semi-supervised learning approaches address many of these issues, these methods assume access to complete datasets, limiting their applicability for real-time use~\cite{yamada2021geoclr,yamada2022guiding}.  
Clustering is an alternative to classification, where data is grouped based on visual similarity without relying on pre-defined class boundaries or label dictionaries. Common approaches include K-means and its variants~\cite{ikotun2023k}, density-based methods like DBSCAN~\cite{bhattacharjee2021survey} and Bayesian clustering~\cite{wu2022modeling}, all of which identify intrinsic patterns within a dataset's latent representation space, where features can be reliably extracted using pretrained visual transformers~\cite{dino} or CNN-based encoders~\cite{yamada2021geoclr}. Studies have shown that these methods can achieve strong value alignment with human class boundaries when used offline~\cite{yamada2021learning}. However, grouping similar scenes in real-time as data is gathered introduces additional challenges~\cite{kudithipudi2022biological, sun2021and}:

\begin{itemize}
    \item Patterns need to be efficiently identified within real-time processing and memory constraints as data volumes grow. 
    \item Cluster boundaries need to adapt in order to integrate new data while avoiding unbounded growth in the number of clusters or compromising the description of existing patterns.
    \item Clustering results should be consistent over accumulated datasets regardless of the order in which observations are made.
\end{itemize}

\noindent Fig.~\ref{fig:method_overview} presents an overview of the method developed in this work. The key contribution is the automated dynamic splitting and merging of clusters in order to maintain a set of clusters that is appropriate over the full history of observations. A key aspect lies in the automatic identification of a subset of representative images that captures the range of features seen, allowing patterns to be distinguished or grouped without the need to accessing the entire, growing history of data. Since this subset of representative images does not grow with data volume, our approach maintains time-constant performance as the data volume accumulates. The approach keeps the number of clusters bounded through efficient merging operations carried out throughout the data collection and performs better than random sub-sampling approaches when class-imbalance exists in datasets, where minor semantic groups are prone to being missed.

\section{Literature review}
 
Online, or continuous, learning involves the development of models that can adapt to new tasks without compromising performance on previously learned ones, all while operating within memory and computational constraints as tasks and data accumulate over time~\cite{wu2019large,rao2019continual}. In clustering, this requires continuous updates to the boundaries that delineate a latent representation space so that new observations are grouped with semantically similar instances, taking into account the entire history of observations. Such updates are necessary when new observations represent previously unseen semantic instances, or reveal the need for finer granularity within existing groupings, or indicate that initially separated clusters lie along a broader continuum of patterns. 

A central challenge in online clustering is balancing the effectiveness of models on new and old data~\cite{rolnick2019experience,yang2022lifelong}. Naively prioritising recent data can lead to rapid degradation in performance on earlier data~\cite{hu2013incremental}, while prioritising historical data can limit the model’s ability to detect patterns in new data as the proportion of historical data grows~\cite{chaudhry2019tiny,kim2019incremental}. Furthermore, since online approaches often need to work in real-time and for long periods without slowing down, it is necessary to retain previous knowledge without re-processing an ever growing volume of past observations.

Knowledge distillation attempts to strike this balance using student-teacher frameworks when training neural networks~\cite{wu2019large,gou2021knowledge}. Teacher models are designed to be more stable and exhibit strong generalisation over past data. These are used to guide the representations or predictions of student models, which are designed to be more adaptable to new data instances. In 
~\cite{hinton2015distilling,asif2019ensemble}, teacher models were pre-trained and used to train a student model online. In~\cite{kim2021feature}, both the teacher and student models were trained online as new data became available, but used different learning rates. Data replay is also deployed to keep historical knowledge. In \cite{tasar2019incremental}, a model was continually trained on new data, but previous versions of it were also stored, and their predictions were used in a training loss function. This encouraged the model to retain its past performance.

Since training neural networks can be computationally expensive, other classical approaches are investigated to achieve online clustering. Bayesian learning method based on Dirichlet Process Mixture Models (DPMM) has been demonstrated to both adapt clusters and create new ones as data arrives~\cite{yang2022lifelong}. Rather than re-training models on the entire historical dataset, their method processes data in batches and incrementally updates the model by combining cluster summary statistics generated for new batches with those from earlier ones. However, their approach cannot adjust existing clusters, which can be limiting if new data reveal subgroups within existing clusters. 


In~\cite{lughofer2015autonomous}, dynamic split–merge operations driven by cluster homogeneity and a penalised Bayesian information criterion BIC allow models to evolve without proliferating clusters. A central issue is deciding when incoming data are sufficiently novel to warrant a new cluster. In \cite{rao2019continual}, samples with low log-likelihood under existing clusters were buffered; a new cluster was created once the buffer exceeded a size threshold at the risk of mixing semantically distinct samples. To mitigate this, \cite{yang2022lifelong} first clustered the provisional buffer before instantiating new clusters. To curb growth, overlapping or similar clusters must be merged. In \cite{lughofer2015autonomous}, merging is guided by the Covariance Hyperellipsoid Volume (CHV): clusters with nearby centres are merged only if the merged CHV volume is approximately no larger than the sum of the individual CHVs; otherwise, merging is discouraged.

Cluster splitting can be framed as an MCMC search over split proposals \cite{dinari2022sampling}, where candidates are iteratively sampled and scored via the Hastings ratio comparing the posteriors of split versus unsplit models; deep Bayesian variants also adopt the same idea \cite{ronen2022deepdpm}. The drawback is the large number of iterations required, making MCMC costly on big datasets. Deterministic surrogates assess splits using latent-space intra-cluster distances \cite{ding2002cluster}. Information criteria offer model-based tests: BIC balances fit and complexity and becomes more conservative as dataset size grows, whereas Akaike information criterion (AIC) applies a linear penalty in the number of clusters, preserving adaptability at scale \cite{wagenmakers2004aic}. Alternatively, geometric heuristics split along principal axes \cite{lughofer2015autonomous} or across multiple dimensions \cite{yang2022lifelong,hwang2019auv}, thereby reconfiguring component structure and steering future sample assignments.

The above methods enable effective clustering of growing datasets but largely ignore how to integrate new observations with an expanding history under fixed memory and compute. Online learning typically relies on replay to preserve knowledge: generate pseudo-samples approximating past data \cite{creswell2018generative} or retain a subset of real samples \cite{yu2023dataset}. For example, \cite{rao2019continual} fuses pre-trained generative model to produce samples with new data during continual representation learning, avoiding full historical storage but requiring sufficiently faithful generators. Alternatively, \cite{lin2022anchor} replays a compact set of representative samples to curb forgetting and cost, but the subset must match the historical latent distribution—random sampling can miss minority classes. Approaches for the representative selection are comprehensively reviewed in \cite{sachdeva2023data,lei2023comprehensive}. Domain-specific schemes include the local pivotal method for spatially heterogeneous geodata \cite{grafstrom2012spatially,dubin1992spatial}; statistical selection of $<0.01\%$ of $1.22\times10^{9}$ satellite pixels to estimate confidence intervals and entropy \cite{blatchford2021determining}; and HK-means representatives that cover cluster centres and boundaries to improve class-imbalanced learning \cite{yamada2022guiding,yamada2021geoclr}, outperforming random or naive stratification \cite{daszykowski2002representative}. However, these studies are offline with full datasets; their efficacy in online setting remains open.

\section{Method}

The OCF maintains a cluster set $\mathcal{S}$ for incremental assignment (Fig.~\ref{fig:method_overview}). New samples are embedded with a pretrained encoder; once a batch threshold is reached, the batch is clustered and its latents are stored. If historical clusters exist, the new clusters are merged using centroid separation and the CHV criterion~\cite{lughofer2015autonomous}. We then pool new and historical latents and select a small representative subset (4k) for splitting: latent-space densities are recomputed (updating only points affected by the new batch), and representatives are stratified by density with the subset size limited by hardware capabilities. Cluster splitting is decided by an information criterion. Because the representative set size is fixed, the effective BIC penalty behaves similarly to AIC. In addition, the OCF adopts a DPGMM backbone, a nonparametric Bayesian mixture, initialised on prior data and updated online. Merge and split operations retrieve sufficient statistics and update the model by parameter reassignment.

\subsection{Dirichlet Process Gaussian Mixture Models}
Dirichlet Process Gaussian Mixture Models (DPGMM), denoted as $DP(\alpha,H)$, are non-parametric models that can be applied to clustering problems. DPGMM can be expressed as (\ref{eq:dpgmm}):
\begin{align}
\phi | \alpha &\sim \text{GEM}(\alpha)\\
z_i | \phi &\sim \text{Mult}(\phi)\\
\theta_k^* | H &\sim H\\
x_i | z_i, \{\theta_k^*\} &\sim F(\theta_{z_i}^*)
\label{eq:dpgmm}
\end{align}
where the $GEM$ (Griffiths-Engen-McCloskey) distribution is the stick-breaking process to generate the weights~\cite{broderick2012beta} and $\alpha$ is the scale parameter coming from $\Gamma$ distribution. $z_i$ is the cluster assignment variable, which takes on the value $k$ with probability $\phi_k$. $z_i$ can be seen as being drawn from a multinomial distribution (Mult($\cdot$)) with the probability vector $\phi$. Because $\phi_k$ decreases exponentially with increasing $k$, only a small number of clusters will be used to model the data.

\subsection{Online Clustering Framework}

\subsubsection{Cluster Merging}
Cluster merging is essential to avoid the explosion of the number of clusters and over-fragmentation of the clustering results. When the latest batch of data $D_i$ is observed, they are first clustered by a $K$-means. Clusters that are in close proximity to existing ones will be merged, while the remaining clusters are treated as newly formed and added to the clustering model. Merging similar clusters helps reduce redundancy~\cite{lughofer2012dynamic}, thereby preventing unnecessary complexity and slowing down the downstream decision-making. In the proposed OCF, we utilise two standards to ascertain if two clusters can be merged:

(1) \textit{Distance criterion} The covariance ellipsoids (for three-dimensional latent spaces; or hyperellipsoids for higher-dimensional latent spaces) of two merging candidates shall be overlapping. Instead of measuring the distance of two clusters in each dimension~\cite{lughofer2012dynamic}, we deploy the Mahalanobis distance to determine whether this criterion is fulfilled. Assuming that $\boldsymbol{\mu}_1$, $\boldsymbol{\mu}_2$ are mean values and $\boldsymbol{\Sigma}_1$ and $\boldsymbol{\Sigma}_2$ are covariances of two clusters, the Mahalanobis distance $D_M$ is determined as (\ref{eq:mahalanobis}):
\begin{equation}
    D_M(\boldsymbol{\mu}_1,\boldsymbol{\mu}_2) = \sqrt{(\boldsymbol{\mu}_1-\boldsymbol{\mu}_2)^T \bar{\boldsymbol{\Sigma}}(\boldsymbol{\mu}_1-\boldsymbol{\mu}_2)^T}
    \label{eq:mahalanobis}
\end{equation}
where $\bar{\boldsymbol{\Sigma}}$ is the average of $\boldsymbol{\Sigma}_1$ and $\boldsymbol{\Sigma}_2$.

If the $D_M(\boldsymbol{\mu}_1,\boldsymbol{\mu}_2)<\varepsilon_{D}$, where $\varepsilon_{D}$ is a calibration parameter, which determines the minimum required overlap, these two clusters are considered eligible for merging. By increasing $\varepsilon_{D}$, more clusters can be merged; otherwise, fewer merging candidates will be available. It is recommended to set $\varepsilon_{D}$ between 3 and 5, as this range corresponds to 99\%-99.999\% confidence regions defined by the chi-square distribution in $D$ dimensions. Setting the $\varepsilon_{D}$ between 3 and 5 provides a practical trade-off: it avoids excessive fragmentation while preventing over-merging of distinct clusters, balancing cluster granularity with statistical consistency in $D$-dimensional space.

(2) \textit{Volume criterion} 
If the clustering merging is performed, the distribution of the merged cluster shall align closely with the original data distribution. The volume criterion is defined as (\ref{eq:merging_criteria}):
\begin{equation}
    \frac{\mathcal{V}_{merged}}{\mathcal{V}_1 + \mathcal{V}_2} \le \varepsilon_V
    \label{eq:merging_criteria}
\end{equation}
where $\mathcal{V}_1$, $\mathcal{V}_2$ and $\mathcal{V}_{merged}$ denote the volumes of two initial and the merged clusters. The $\varepsilon_V$ is a calibration parameter that is typically set close to 1.0 to avoid over-merging.

The volume of a cluster is calculated as:
\begin{equation}
    \mathcal{V} = \frac{\pi^{d/2}}{\Gamma(d/2)} \times \prod_{i=1}^{d} \sqrt{\lambda_i} \times \chi^2_{0.95}(d)^{d/2}
\end{equation}
where $\lambda_i$ is the $i$-th eigenvalue of the covariance matrix $\boldsymbol{\Sigma}$, $\chi^2_{0.95}(d)$ is the 95\% quantile of the chi-squared distribution with $d$ degrees of freedom and $\Gamma$ is the Gamma distribution.

When determining whether the volume condition is satisfied, it is essential to calculate the centre $\boldsymbol{\mu}_{merged}$ and the covariance $\boldsymbol{\Sigma}_{merged}$ of the merged clusters. Given the weights $\phi_1$ and $\phi_2$ of two separate clusters, 
\begin{align}
    \boldsymbol{\mu}_{\text{merged}} &= \frac{\phi_1 \boldsymbol{\mu}_1 + \phi_2 \boldsymbol{\mu}_2}{\phi_1 + \phi_2}, \\
    \boldsymbol{\Sigma}_{\text{merged}} &= \large\{\phi_1 (\boldsymbol{\Sigma}_1 + (\boldsymbol{\mu}_1 - \boldsymbol{\mu}_{\text{merged}})(\boldsymbol{\mu}_1 - \boldsymbol{\mu}_{\text{merged}})^T)\notag\\
    &+ \phi_2 (\boldsymbol{\Sigma}_2 + (\boldsymbol{\mu}_2 - \boldsymbol{\mu}_{\text{merged}})(\boldsymbol{\mu}_2 - \boldsymbol{\mu}_{\text{merged}})^T)\large\}\\
    &/(\phi_1 + \phi_2)~. \notag
\end{align}

Based on the volume criteria, a substantial increase in the volume of a merged cluster compared to the sum of volumes of the individual clusters indicates reduced homogeneity. In such a case, merging these clusters is not desirable. Cluster merging is performed recursively to make sure that there are no mergeable cluster candidates left. Then, the backbone model $\mathcal{S}$ is updated:
\begin{align}
    \mathcal{S} &\gets \mathcal{S} - \mathcal{S}_{mergeable}\\
    \mathcal{S}_{merged} &\gets \Large[\phi_{merged}, \boldsymbol{\mu}_{merged}, \boldsymbol{\Sigma}_{merged}]\\
    \mathcal{S} &\gets \mathcal{S} + \mathcal{S}_{merged}
\end{align}
where $\mathcal{S}_{mergeable}$ is the components of mergeable clusters which are removed from the model. Instead, the $\mathcal{S}_{merged}$ will be integrated into the clustering model.

\subsubsection{Cluster Splitting}

An existing cluster may need to be split into two or more clusters when more data is observed. However, performing cluster splitting presents significant challenges, particularly in the OCF, where only summary statistics of the observed data are maintained. To address this problem, we leverage the representative samples through data distillation, which are discussed later, to model the latent distribution of all history data and update the clustering model. 

The splitting operation is conducted recursively. In each operation, we deploy Gaussian Mixture Models (GMM), as noted in (\ref{eq:gmm}), to determine whether the existing clusters should be further split. 

\begin{equation}
p(\mathbf{x}) = \sum_{k=1}^{K} \pi_k \cdot \mathcal{N}(\mathbf{x} \mid \boldsymbol{\mu}_k, \boldsymbol{\Sigma}_k)
\label{eq:gmm}
\end{equation}
where $\pi_k$ is the weight of component and $\mathcal{N}(\cdot)$ is the Gaussian distribution. 

Specifically, each cluster is split into two sub-clusters ($k=2$), and their statistical description is evaluated using a criterion. If the total information gain is significant, the split is accepted. Otherwise, the original cluster is retained. In this study, we can alternatively deploy two splitting criteria, namely AIC and BIC. 

\begin{equation}
    AIC = 2 \cdot k - 2 \cdot \log (\hat{\mathcal{L}})
\end{equation}

\begin{equation}
    BIC = -2 \cdot \log (\hat{\mathcal{L}}) + k \cdot \log (N)
    \label{eq:bic}
\end{equation}
where $\hat{\mathcal{L}}$ is the maximised likelihood, $N$ is the number of observed data and $k$ denotes the number of parameters in the model.

Assuming the collected data follows the Gaussian distribution, the maximised likelihood can be expressed as:

\begin{equation}
    \log (\hat{\mathcal{L}}) = -\frac{n}{2} \log(2\pi \sigma^2)- \frac{n}{2 \sigma^2} \sum_{i=1}^{n}(x_i - \hat{x}_i)^2
\end{equation}
where $x_i$ and $\hat{x}_i$ are the sample data $i$ and its predictions.

The model with a lower AIC/BIC can be seen as better than that with a higher AIC/BIC. Therefore, if the $\mathcal{M}_1$ has a lower AIC/BIC, the existing cluster is not further split; otherwise, it is. Splitting is performed recursively until no cluster is divisible. Subsequently, new clusters can be used to update the $\mathcal{S}$ in the OCF.

\begin{align}
    \mathcal{S} &\gets \mathcal{S} - \mathcal{S}_{\mathcal{M}_1}\\
    \mathcal{S}_{split} &\gets \Large[\phi_{\mathcal{M}_{2,{1\wedge2}}}, \boldsymbol{\mu}_{\mathcal{M}_{2,{1\wedge2}}}, \boldsymbol{\Sigma}_{\mathcal{M}_{2,{1\wedge2}}}]\\
    \mathcal{S} &\gets \mathcal{S} + \mathcal{S}_{split}~.
\end{align}

\subsubsection{Density-based Data Distillation}\label{data_distillation}

The OCF performs clustering operations only when the newly observed data reach a predetermined count. To prevent a significant slowdown in clustering speed, not all observed samples are used in future clustering. Instead, a small subset is retained to keep representative samples. This subset operates similarly to the queue used in \cite{caron2020unsupervised}, maintaining a small number of image samples during learning. By storing a subset of data, the OCF reduces computational load while capturing the distribution of the entire dataset \cite{yamada2022guiding}.

Assuming that $N_{\text{sub}}$ is the required number of representative samples, it can be pre-defined based on CPU memory constraints. A larger number of representative samples benefits the cluster splitting by providing more comprehensive information, but it also increases computational cost. Conversely, fewer representative samples can accelerate the clustering process, although too few may fail to capture the overall latent distribution of historical data. Therefore, we set $N_{\text{sub}}$ in the range of $200 \times d$ to $300 \times d$, where $d$ is the dimensionality of the latent vectors. In our experiments, we set $d=16$~\cite{liang2025self}, resulting in $N_{\text{sub}} = 4000$. Then, the average Manhattan distance $\bar{\mathcal{D}}$ between each exemplar and its nearest samples $x_i$, where $i=1,2,\cdots, n$ is used to evaluate the local distribution density. A smaller $\bar{\mathcal{D}}$ generally indicates higher distribution density near the sample in feature space. Subsequently, a sorted queue $\pi$ can be obtained:
\begin{equation}
    \bar{\mathcal{D}}_{\pi_{(1)}} < \bar{\mathcal{D}}_{\pi_{(2)}} < \cdots < \bar{\mathcal{D}}_{\pi_{(N)}}  
\end{equation}
where $N$ is the number of observed data.

As the required number of representative samples is $N_S$, we select samples from $\pi$ at the interval of 
\begin{equation}
    m = \left \lfloor \frac{N}{N_{S}} \right \rfloor  > \varepsilon_m
\end{equation}
where $\varepsilon_m$ is a preset minimum value, which ensures there are sufficient samples when the historical dataset is small.

Therefore, the representative samples can be obtained as
\begin{equation}
    \mathbf{x}_r = [x_{\pi_{(1)}}, x_{\pi_{(m+1)}}, x_{\pi_{(2m+1)}}, \cdots ,x_{\pi_{km+1}}  ]
    \label{eq:queue}
\end{equation}
where $km+1 \le N$

When a new batch of data is collected and clustering is triggered, they are added to the historical data, and representatives are generated to facilitate cluster splitting. However, instead of recalculating the local density for all historical data, only existing samples in the vicinity of the latest batch are updated, and the entire queue can be fine-tuned or resorted at a lower frequency.


\section{Experiments and Analysis}

\subsection{Datasets Description} 
\begin{figure*}
    \centering
    \includegraphics[width=0.9\linewidth]{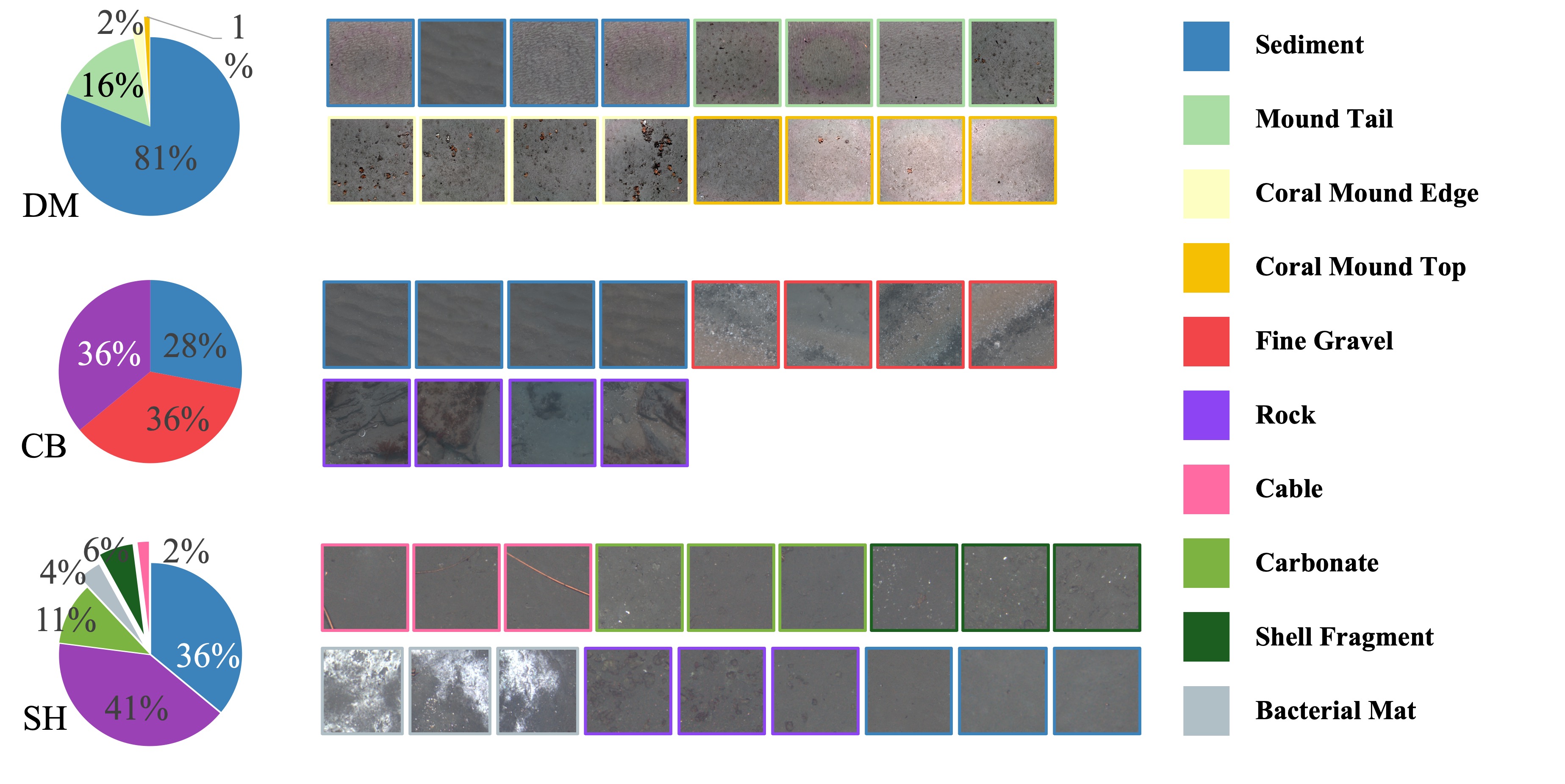}
    \caption{Image examples and class proportions of three field survey datasets used in this paper.}
    \label{fig:representative_images}
\end{figure*}

The Darwin Mounds (DM), Cawsand Bay (CB) and the Southern Hydrate Ridge (SH) seafloor image datasets are used in this paper. They were collected in 2019, 2024 and 2018 by the Autosub6000 (National Oceanography Centre, UK), Smarty200 (University of Southampton, UK) and AE2000f (The University of Tokyo, UK), respectively. These three datasets include around 90,000 seafloor images that are categorised into 10 classes. Class proportions and examples are shown in Fig.~\ref{fig:representative_images}. More information about these three datasets can be found in \cite{Liang2025LRSSL}

\subsection{Experimental setup}

The OCF shown in Fig.~\ref{fig:method_overview} invokes clustering after every 1,000 newly acquired seafloor images. To satisfy onboard CPU limits, the representative set is capped at 4,000 images. Cluster merging uses conservative criteria with distance and volume thresholds $\varepsilon_D=5$ and $\varepsilon_V=1.1$. The model is initialised without priors to avoid bias and adapt solely from incoming data.



Besides, we compare seven clustering variants that differ in backbone, splitting strategy, and representative selection. \textit{Full history} uses a DPGMM backbone and reclusters over all accumulated images (no splitting; no representative selection), whereas all online variants update incrementally on the current batch only. Online methods are: (i) \textit{Only Merging} (DPGMM; merge-only, no splitting; no representatives)~\cite{yang2022lifelong}; (ii) \textit{SAM + Principal Splitting} (GMM; recursive splits along principal variance directions; representatives selected at random)~\cite{lughofer2015autonomous}; (iii) \textit{SAM + Random Sampling} (GMM; splits across multiple dimensions without prioritisation; random representatives); (iv) \textit{SAM + Density-based Sampling} (GMM; multi-dimensional splits; representatives chosen by local-density criterion); (v) \textit{OC + Density-based Sampling} (DPGMM; multi-dimensional splits; density-based representatives); and (vi) \textit{OC + HKmeans Sampling} (DPGMM; multi-dimensional splits; hierarchical k-means for representative selection)~\cite{yamada2022guiding}. 



To evaluate sensitivity to observation order, we employ three survey patterns: (i) a \textit{West–East} raster (lawnmower) with parallel horizontal tracks and U-turns at the boundaries; (ii) a \textit{North–South} raster with parallel vertical tracks; and (iii) a \textit{Random} waypoint pattern composed of short point-to-point transects with arbitrary headings. Consistent results across patterns indicate order robustness and suitability for diverse AUV trajectories, which is required by the practical application of AUVs.


\subsection{Clustering Results of Seafloor Image Datasets}

\subsubsection{Accuracy}

\begin{figure}[!t]
  \centering
  \subfloat{%
    \includegraphics[width=0.9\linewidth]{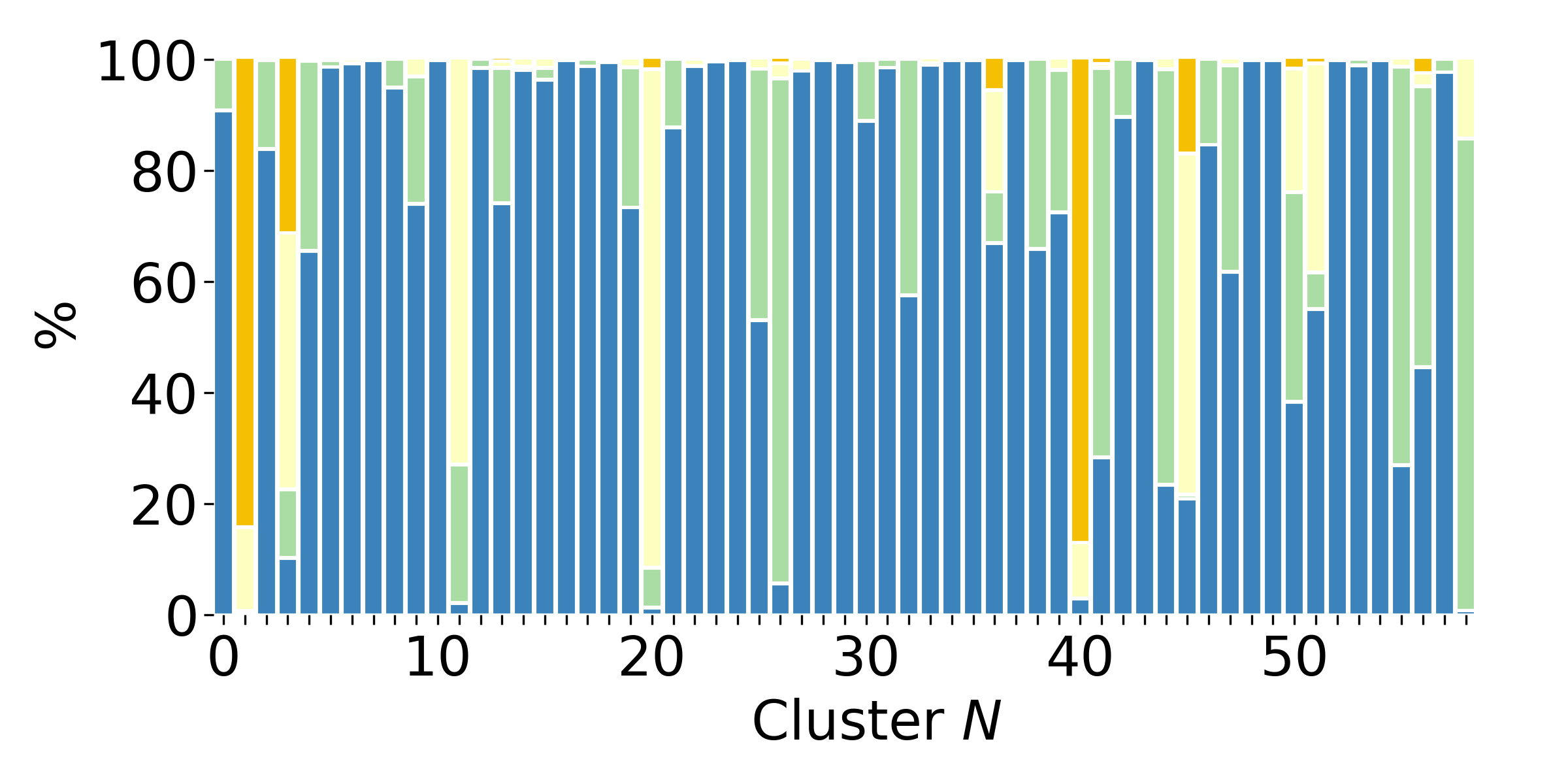}
    \label{fig:stackbar-dm}}
  \\[-1pt]
  \subfloat{%
    \includegraphics[width=0.9\linewidth]{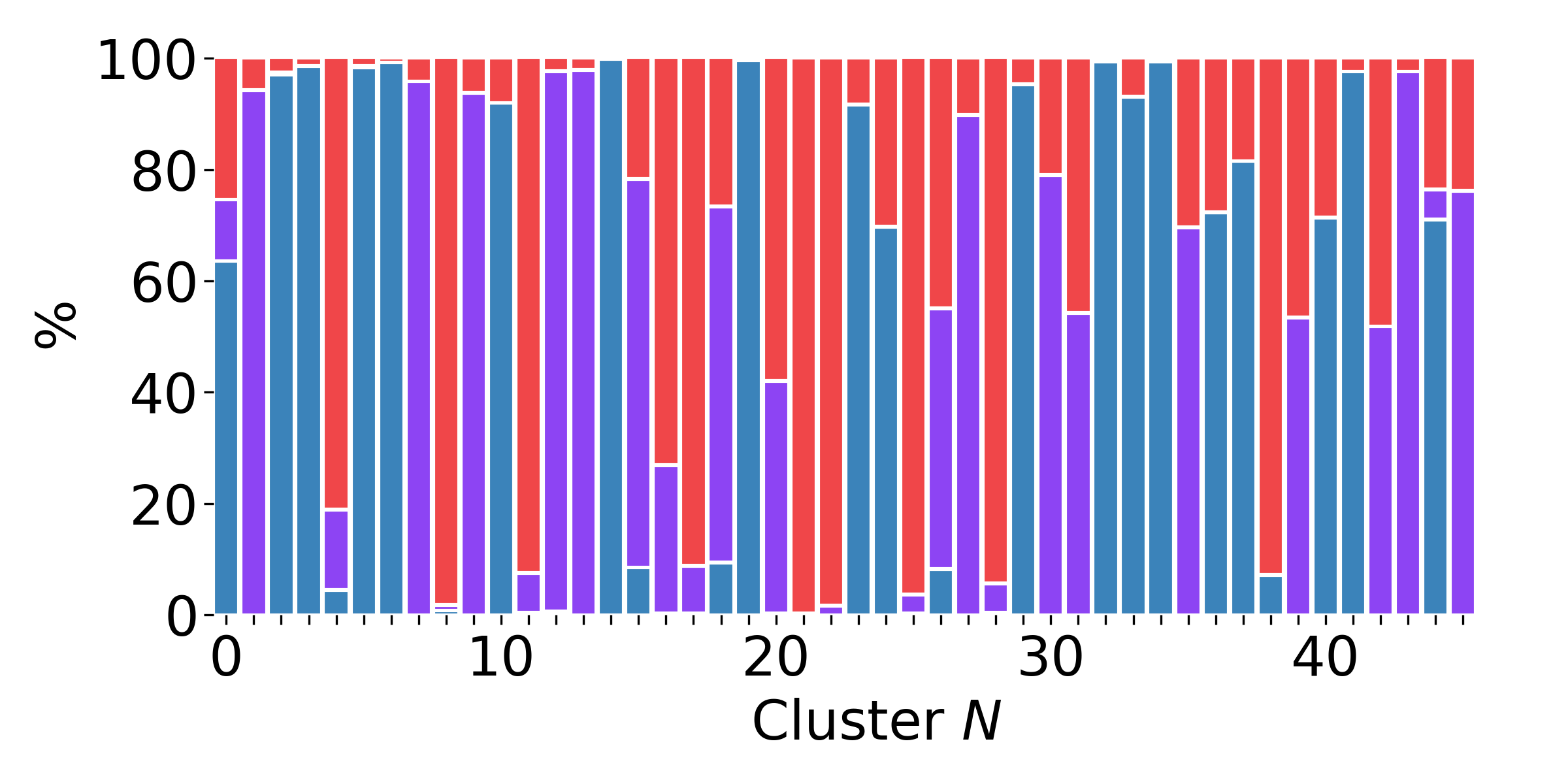}
    \label{fig:stackbar-cb}}
  \\[-1pt]
  \subfloat{%
    \includegraphics[width=0.9\linewidth]{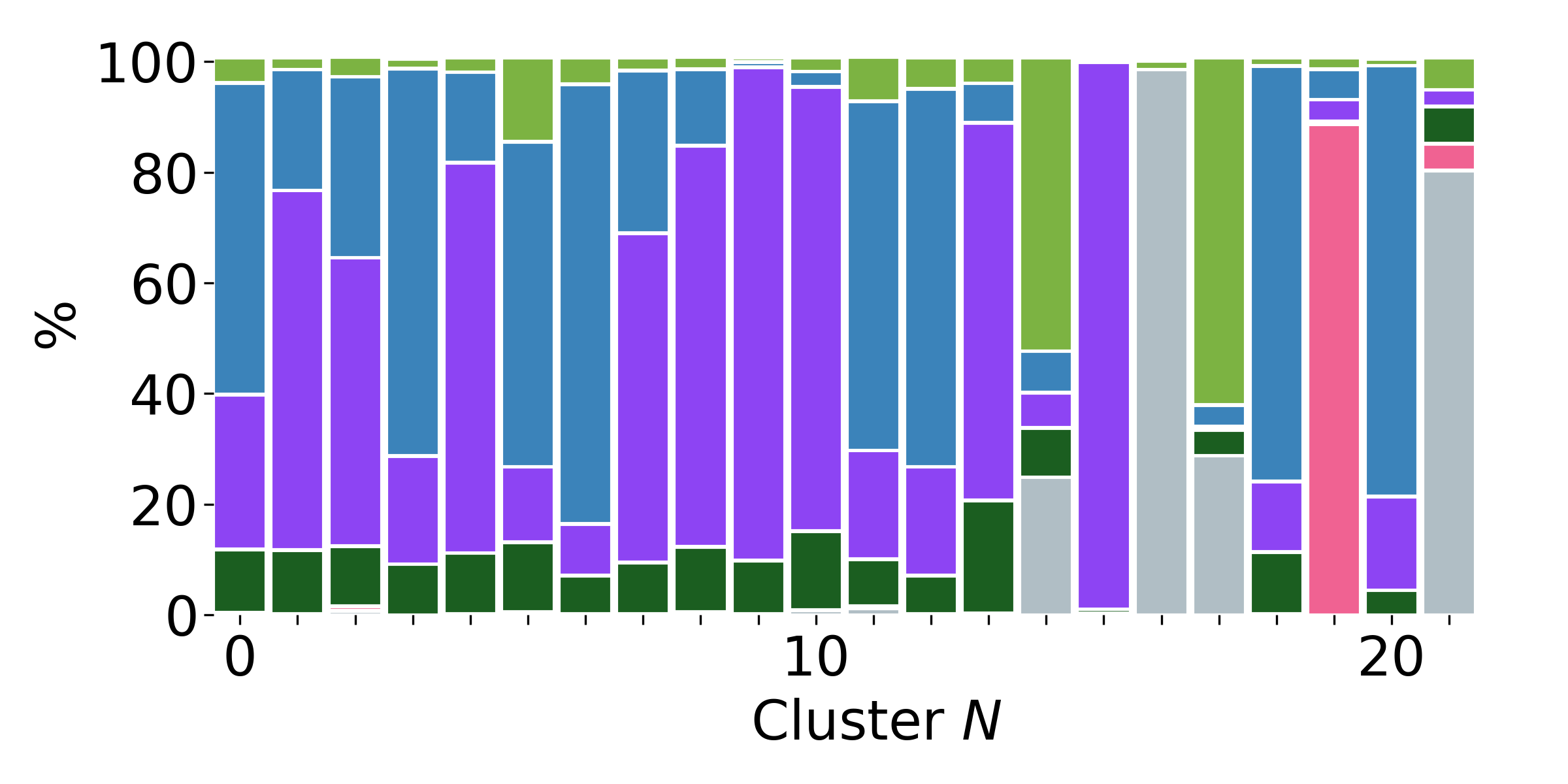}
    \label{fig:stackbar-sh}}
  \caption{Class distributions in each cluster in the DM, CB and SH datasets. Class colours are the same as examples in Fig.~\ref{fig:representative_images}.}
  \label{fig:stackbar}
\end{figure}

The F1 score is employed to assess clustering performance by comparing the predicted cluster assignments with the ground truth class labels. It is important to recognise that the ground truth labels correspond to human-annotated semantic categories, whereas unsupervised clustering organises samples based on feature similarity. As a result, a single cluster may contain samples from multiple semantic classes, as illustrated in Fig.~\ref{fig:stackbar}. This overlap is an inherent characteristic of unsupervised learning, where clusters reflect structure in the feature space rather than strict alignment with predefined class boundaries. Furthermore, the primary objective of our method is not exact class recovery, but rather the grouping of seafloor images according to underlying visual patterns—patterns that may not always align with human-defined categories. To compute the F1 score in such cases, we follow the major-voting strategy~\cite{yamada2021learning} of assigning each cluster the label of its majority class, i.e., the class with the largest number of samples within that cluster.

Fig.~\ref{fig:stackbar} depicts class–cluster relations, revealing inter-class similarity. Under majority voting, 81.0\%, 8.7\%, 6.9\%, and 3.4\% of clusters in the DM dataset are mapped to Sediment, Coral Mound Tail, Coral Mound Edge, and Coral Mound Top, respectively. Relative to ground truth, Tail is occasionally grouped with Edge/Top; given their similar appearances (Fig.~\ref{fig:representative_images}), these overlaps are plausible. In the CB dataset, clusters 20, 31, and 40 mix Rock and Fine Gravel in comparable shares, indicating similar features in transitional habitats. In SH, Shell Fragment is dispersed across many small clusters, suggesting weak cohesiveness or dominance by other patterns. By contrast, the Cable class (2\% of samples) concentrates mainly in cluster 19, showing its significant features.

\begin{figure*}
    \subfloat[]{
        \includegraphics[width=0.33\linewidth]{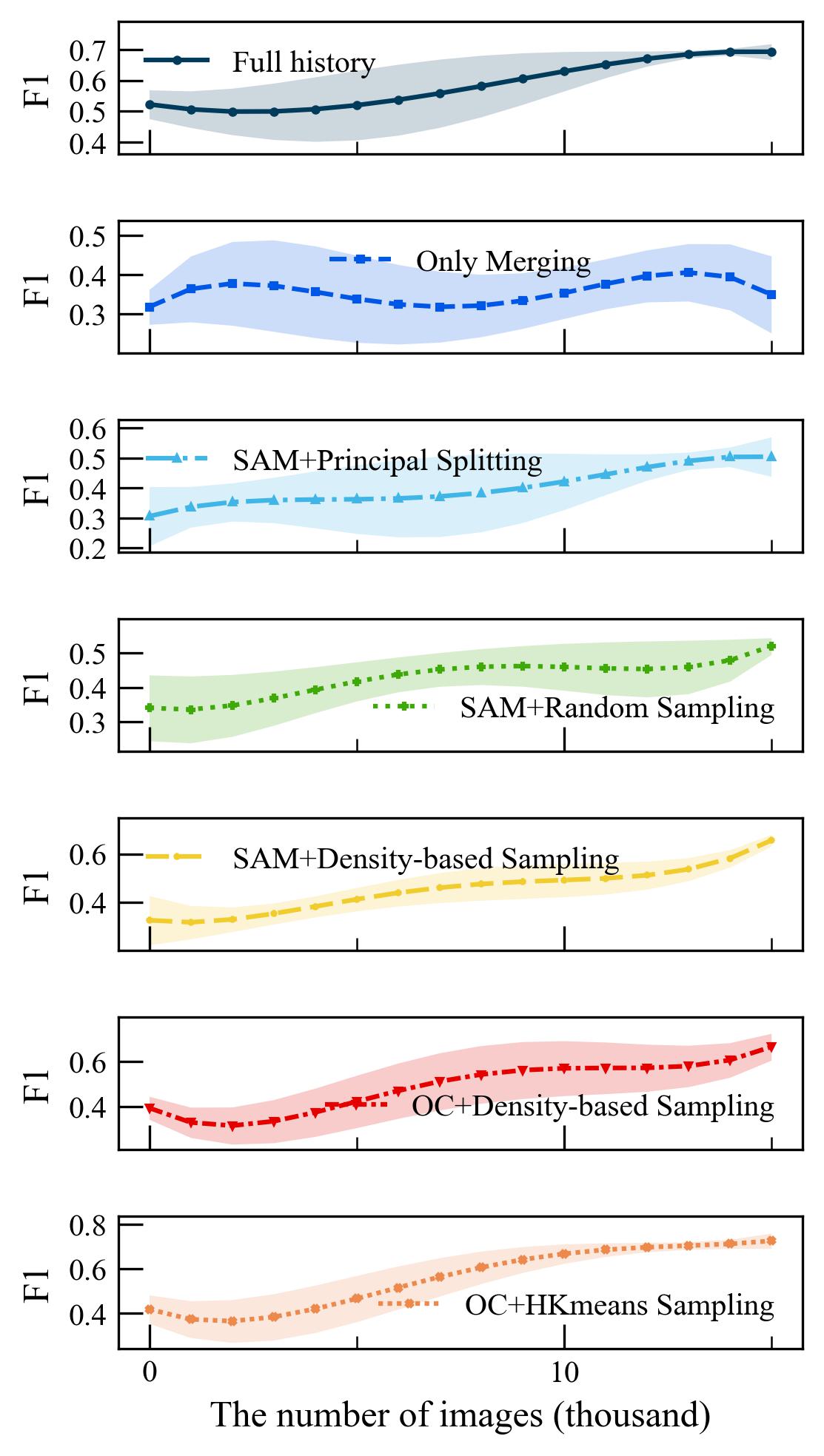}
    }
    \subfloat[]{
        \includegraphics[width=0.33\linewidth]{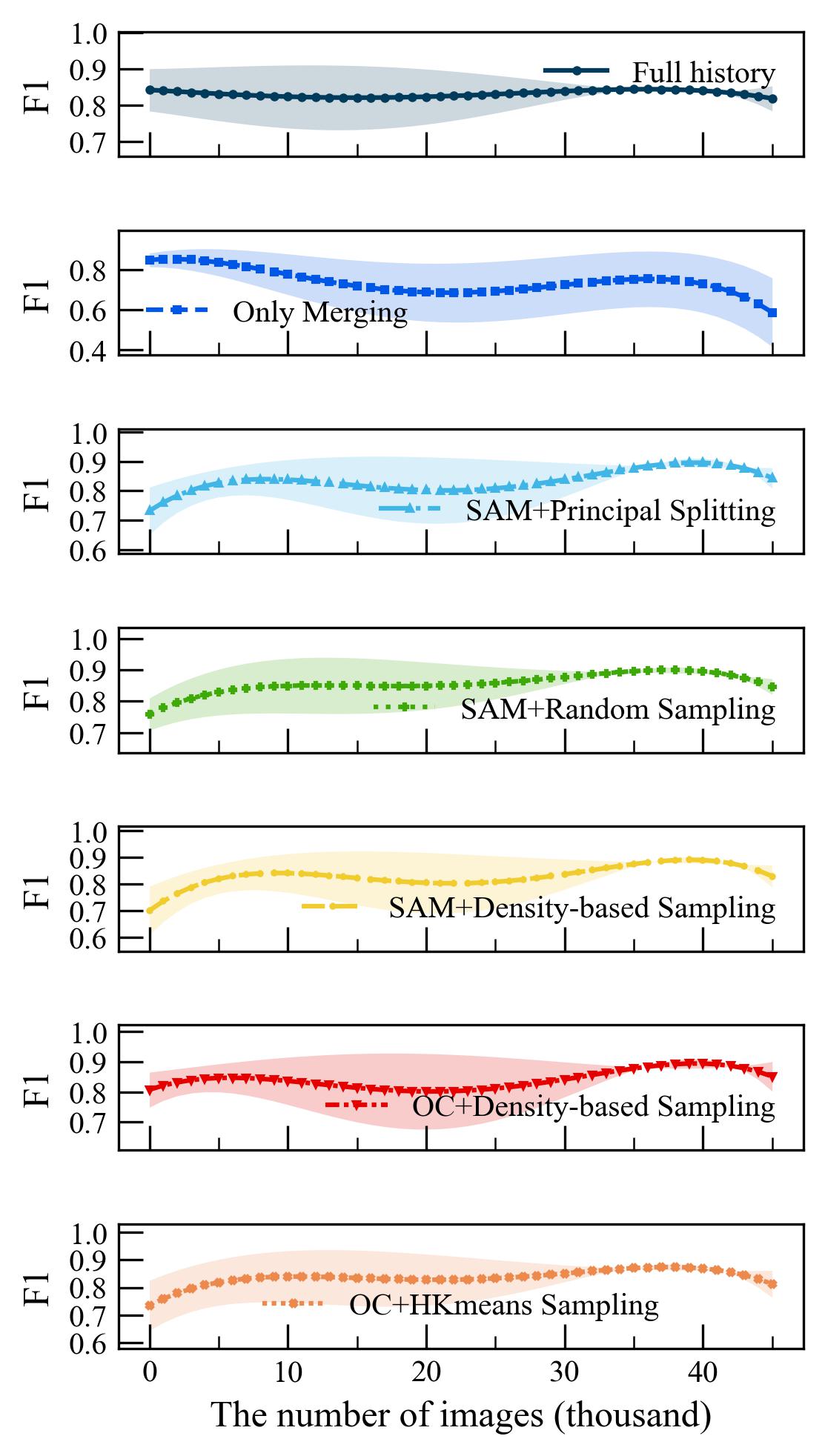}
        }
   \subfloat[]{
        \includegraphics[width=0.33\linewidth]{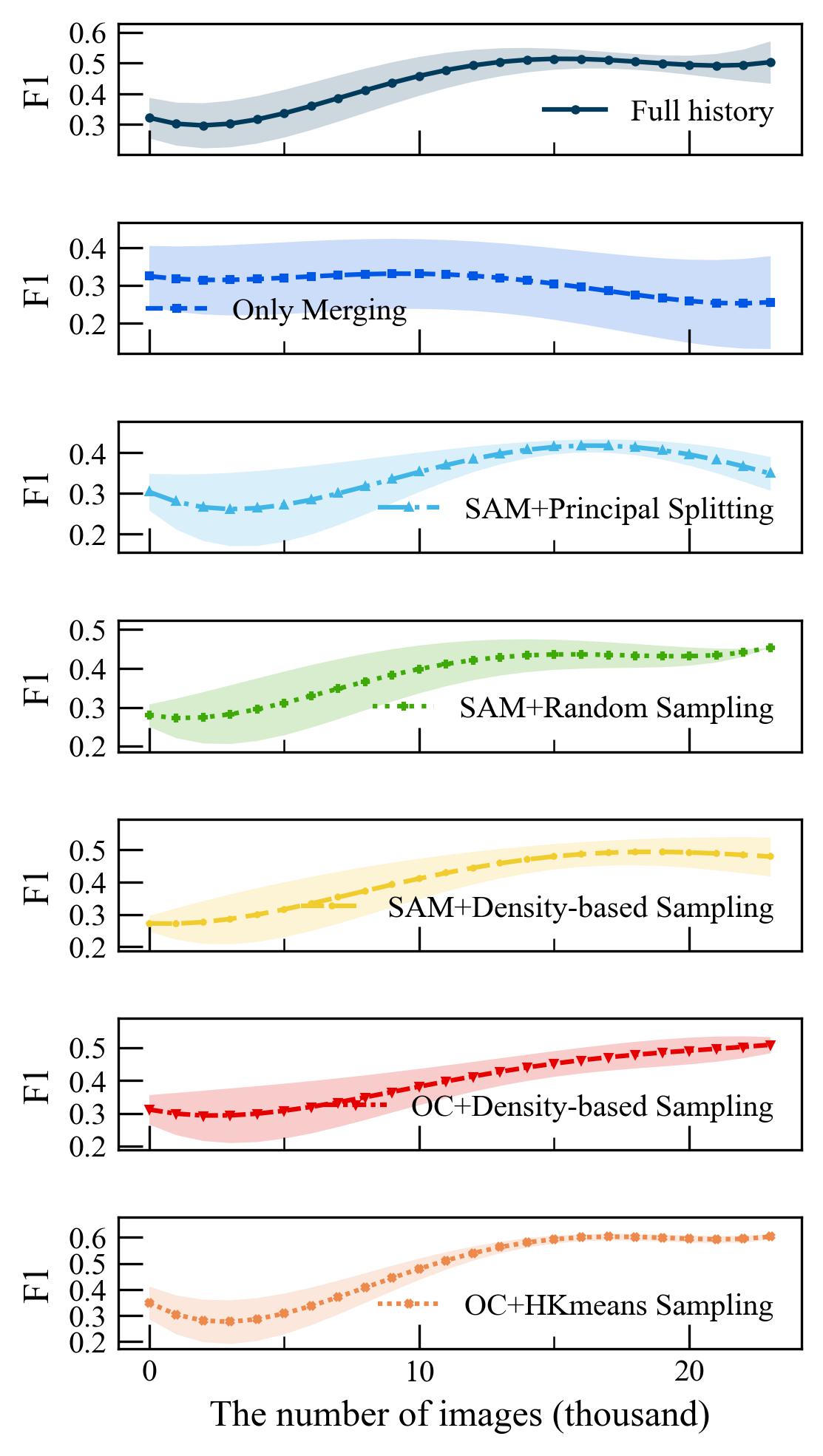}
        }
    \caption{F1 scores of different clustering methods across various seafloor datasets in DM, CB and SH from left to right. Results are smoothed using a 4th-degree polynomial fit for clearer visualisation. Standard deviation is computed based on three independent survey paths.}
    \label{fig:f1_score}
\end{figure*}


Fig.~\ref{fig:f1_score} shows F1 scores across three datasets: (a) DM, (b) CB, (c) SH, as a function of cumulative images. \textit{Only Merging} is consistently worst, underscoring the need for cluster splitting. \textit{OC + HKmeans Sampling} attains the best F1 (0.726), followed by \textit{Full History} (0.701), evidencing the benefit of dynamic merge–split updates over a single-batch static model. Nonetheless, \textit{Full History} remains a strong baseline. The remaining methods rank: \textit{OC + Density-based Sampling} \textgreater {} \textit{SAM + Density-based Sampling} \textgreater {} \textit{SAM + Random Sampling} \textgreater {} \textit{SAM + Principal Splitting}.


Cluster splitting is pivotal because it alters cluster components and parameters that determine the final partition. \textit{SAM + Random Sampling} revises \textit{SAM + Principal Splitting}, which recursively divides along principal axes; while effective, the latter can leave substantial overlap on non-principal directions. By splitting across multiple dimensions and allowing the mixture model to resolve separations in the full feature space, \textit{SAM + Random Sampling} improves F1 by 3.1\% on average across the three datasets.


Representative selection is equally critical, since the splitting estimates component parameters from representatives. \textit{SAM + Density-based Sampling} replaces random selection with the density-guided rule in (\ref{eq:queue}), yielding a further 3.1\% average F1 gain. Random sampling fares well under class balance, e.g., CB attains 0.873, close to density-based, but degrades on imbalanced DM and SH by missing minority clusters. Density-based selection follows local density, providing more even coverage, including minor clusters.


The backbone ultimately governs label inference during online updates. \textit{OC + Density-based Sampling} (DPGMM backbone) attains an average F1 of 0.678, exceeding \textit{SAM + Density-based Sampling} (GMM backbone) by 3.7\%. Although both are initialised with identical mixture weights, means, and covariances, their inference mechanisms differ: GMM assigns labels via maximum likelihood, whereas DPGMM performs Bayesian inference with a Dirichlet-process prior that regularises component usage and penalises unlikely assignments. Consequently, DPGMM is more robust under cluster overlap and data sparsity, producing more stable, conservative posteriors, especially in high-dimensional or ambiguous regions—whereas GMM’s pure MLE tends to be more noise-sensitive and overconfident.


Class imbalance substantially affects clustering. In the CB dataset, where habitat classes are roughly balanced, variations in splitting strategy, representative selection, and backbone have little impact; retrieval of representatives from all classes is likely, and decision boundaries are less skewed. By contrast, DM and SH exhibit pronounced imbalance, making unsupervised clustering difficult: minority classes are underrepresented, especially with random sampling, yielding biased parameter estimates and suboptimal assignments. Imbalance can also distort splitting, as dominant classes overwhelm inference and obscure minority structure. Under these conditions, density-based sampling and nonparametric backbones, i.e., DPGMM, better preserve minority classes by focusing on local density and allowing flexible cluster complexity, thereby mitigating imbalance-induced bias.

\subsubsection{Time Cost}

\begin{figure*}
    \subfloat[The DM dataset]{
        \includegraphics[width=0.33\linewidth]{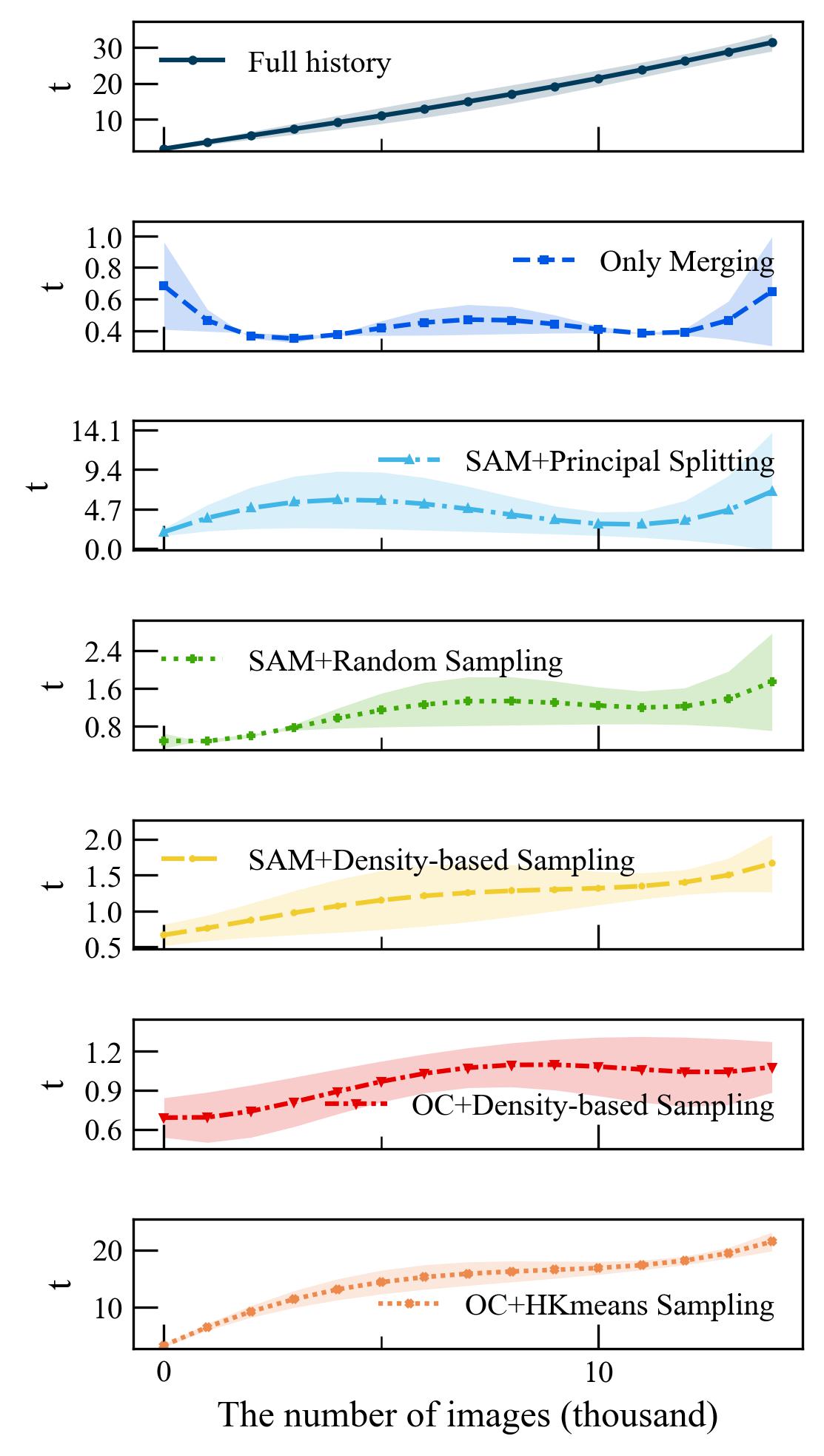}
    }
    \subfloat[The CB dataset]{
        \includegraphics[width=0.33\linewidth]{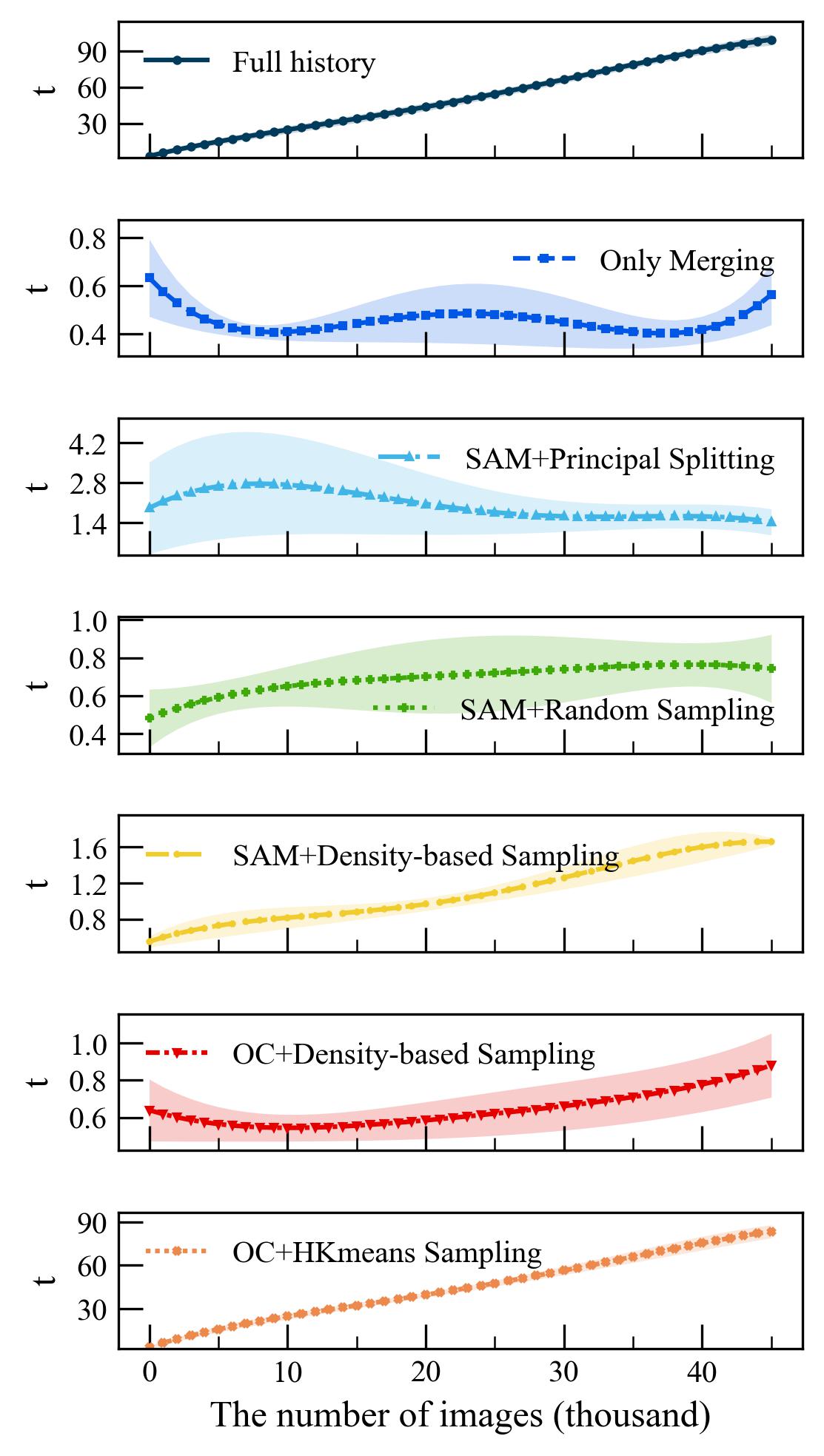}
        }
   \subfloat[The SH dataset]{
        \includegraphics[width=0.33\linewidth]{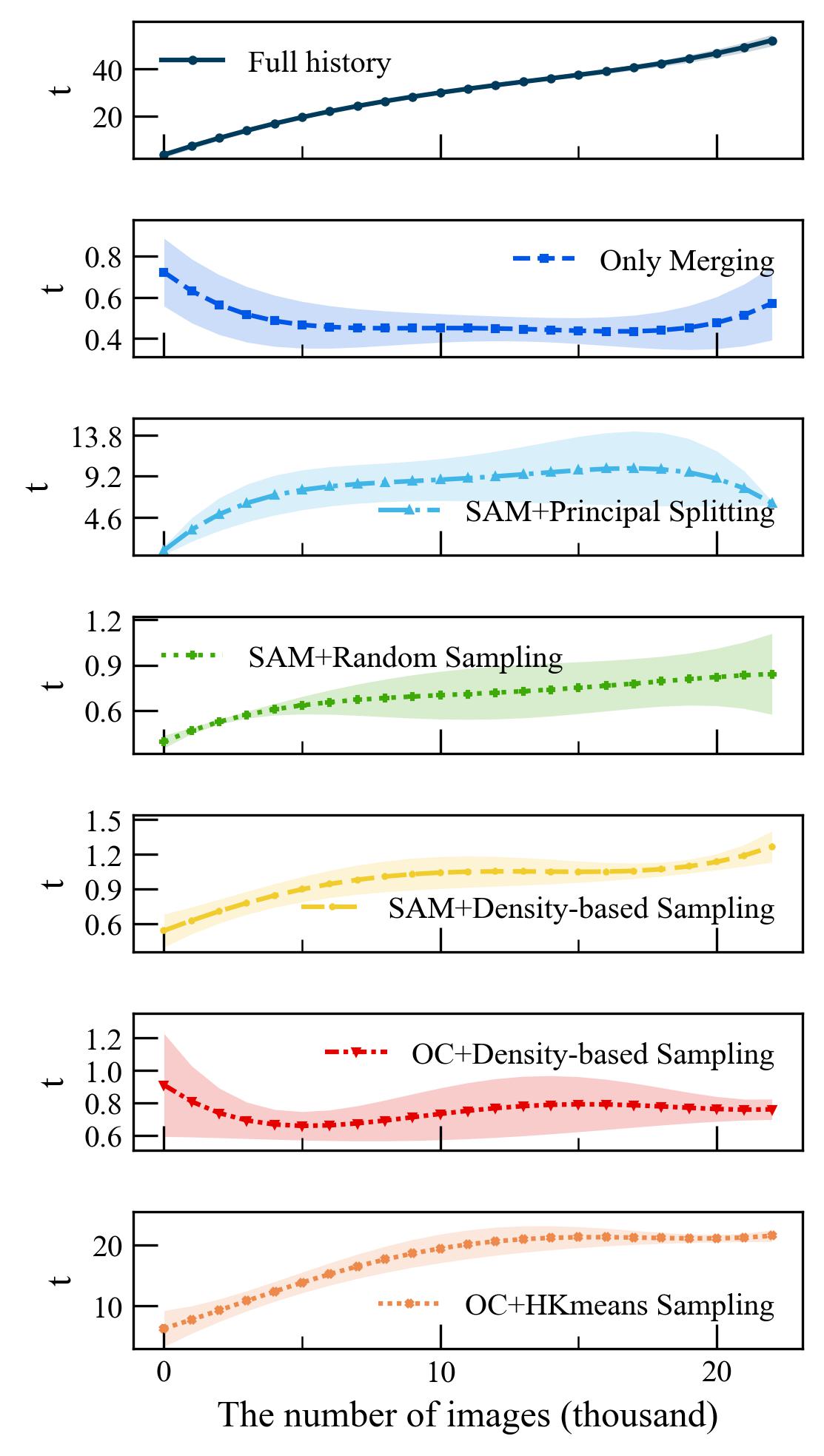}
        }
    \caption{Time cost of different clustering methods across various seafloor datasets. Results are smoothed using a 4th-degree polynomial fit for clearer visualisation. Standard deviation is computed based on three independent survey paths.}
    \label{fig:time_cost}
\end{figure*}

Fig.~\ref{fig:time_cost} illustrates per-method computation time during online clustering; the x-axis is cumulative images (k) and the y-axis is seconds. \textit{Full History} and \textit{OC + HKmeans Sampling} incur the highest costs because they replay all past images at each update. For example, in Fig.~\ref{fig:time_cost}(a) \textit{Full History} takes around 30 s at 18k images and \textgreater{100} s at 40k, undermining real-/near-real-time use in long missions. \textit{SAM + Principal Splitting} is third in time but bounded; its cost scales with feature dimensionality due to recursive principal-axis checks. \textit{Only Merging}, \textit{SAM + Random/Density Sampling}, and \textit{OC + Density} complete in a few seconds, satisfying online requirements. Balancing accuracy and latency, \textit{OC + HKmeans Sampling} and \textit{Full History} exceed 0.7 F1 across datasets but scale poorly in time, whereas \textit{OC + Density-based Sampling} offers a better trade-off: density-guided representative selection cuts runtime relative to HKmeans while outperforming random sampling in clustering quality.


\subsubsection{Inferred Clustering Results}

\begin{figure}
    \subfloat[Ground truth]{
        \includegraphics[width=0.3\linewidth]{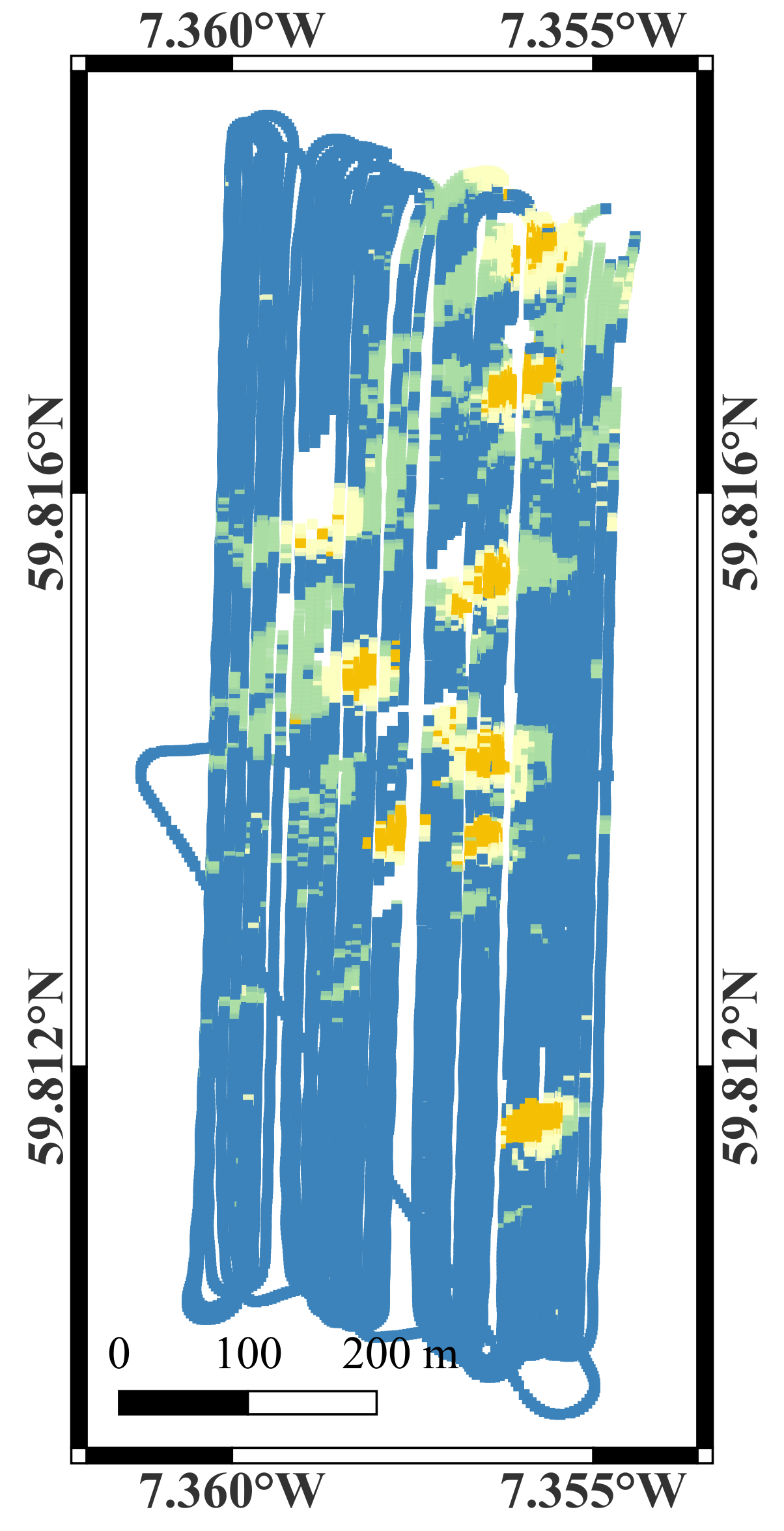}
    }
    \subfloat[Consistency map]{
        \includegraphics[width=0.3\linewidth]{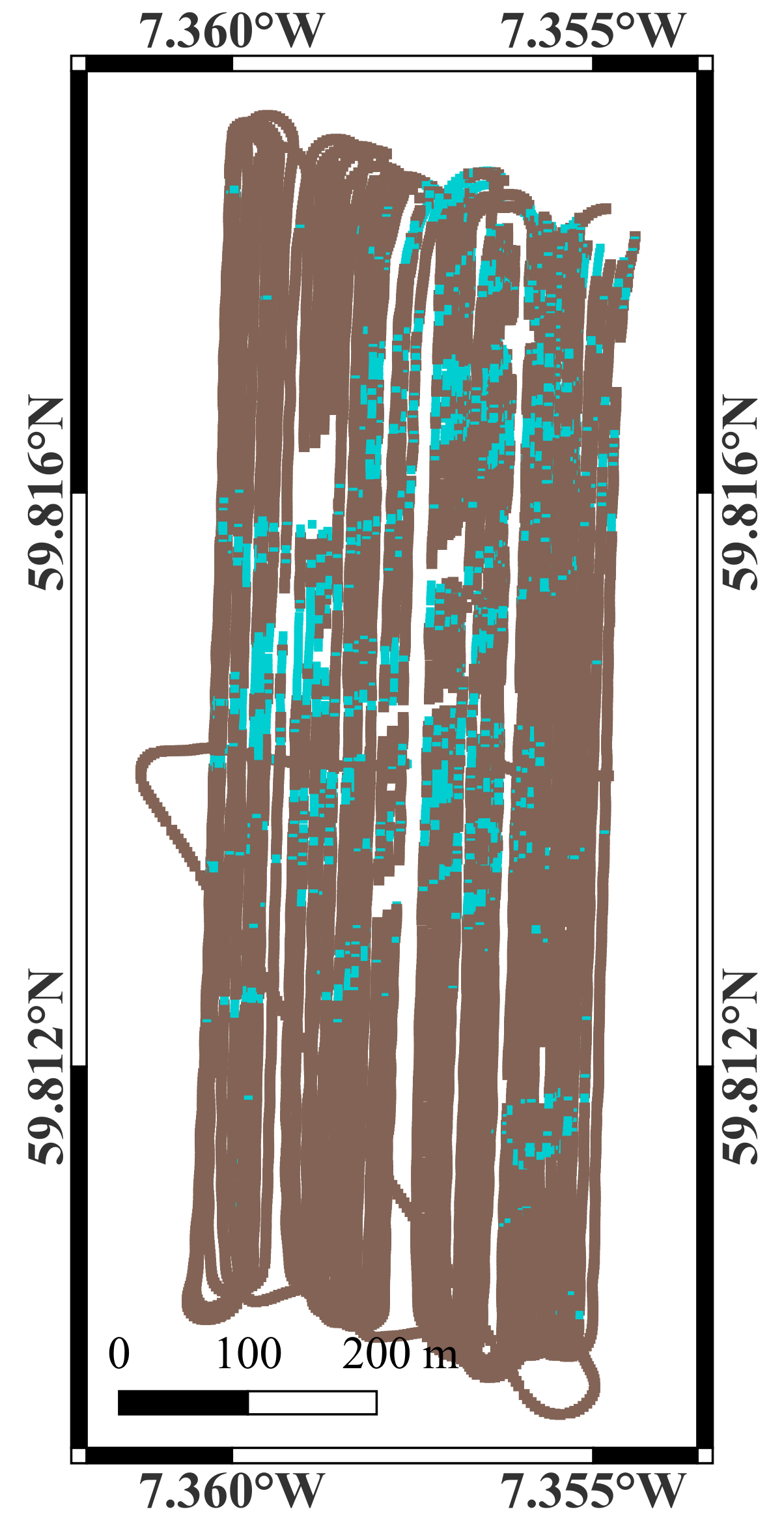}}
    \subfloat[Entropy map]{
        \includegraphics[width=0.3\linewidth]{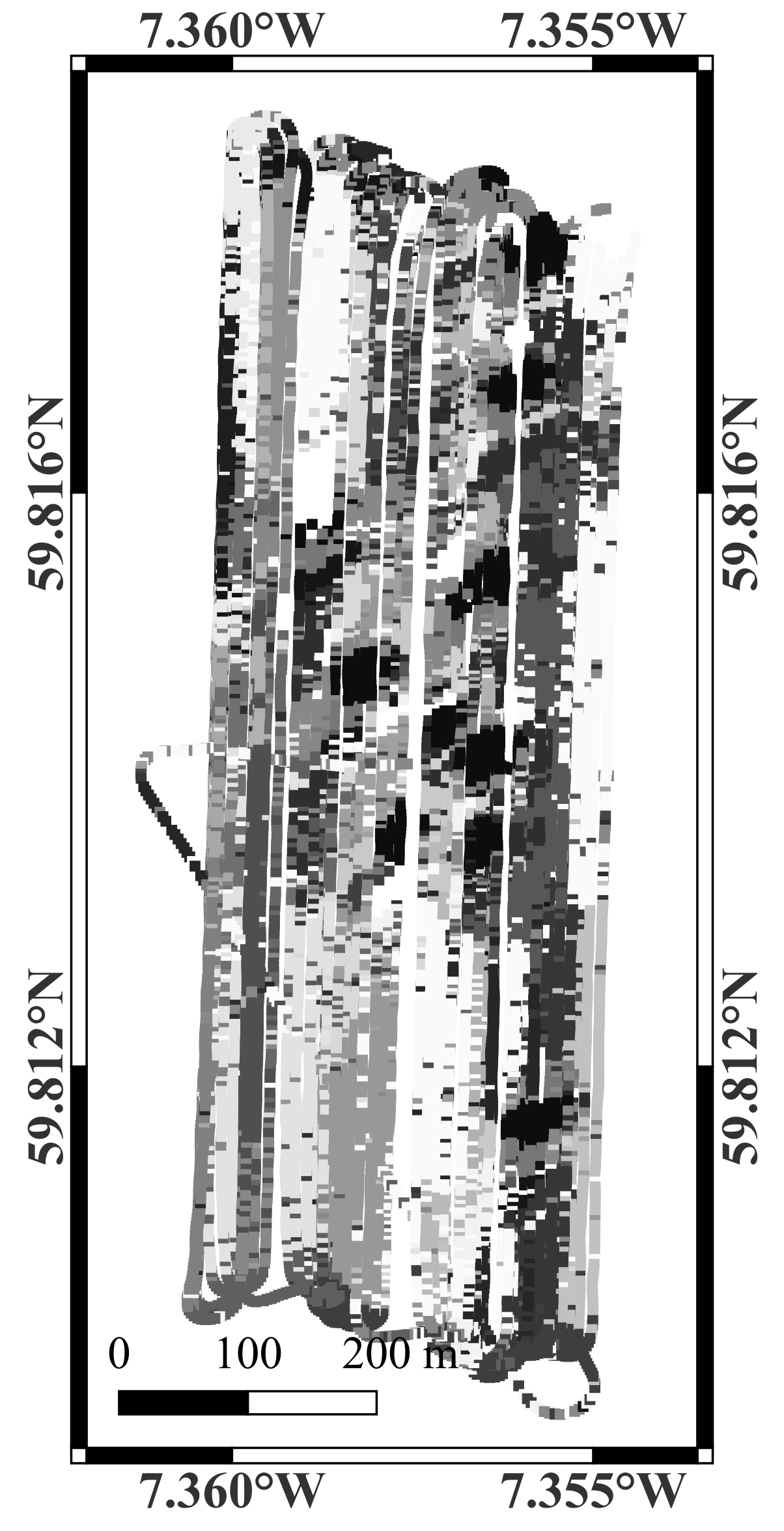}}
        
    \subfloat{\includegraphics[width=\linewidth]{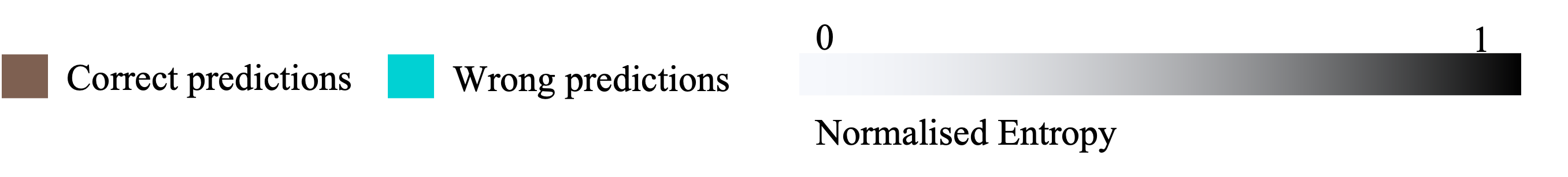}}
    \caption{Comparison of the ground truth and the inferred clustering results from the online clustering for the DM dataset. Panel (c) displays the entropy of each inferred cluster, which visualises the purity of the ground truth class distribution within that cluster.}
    \label{fig:dm_result}
\end{figure}

\begin{figure}
    \subfloat[Ground truth]{
        \includegraphics[width=0.3\linewidth]{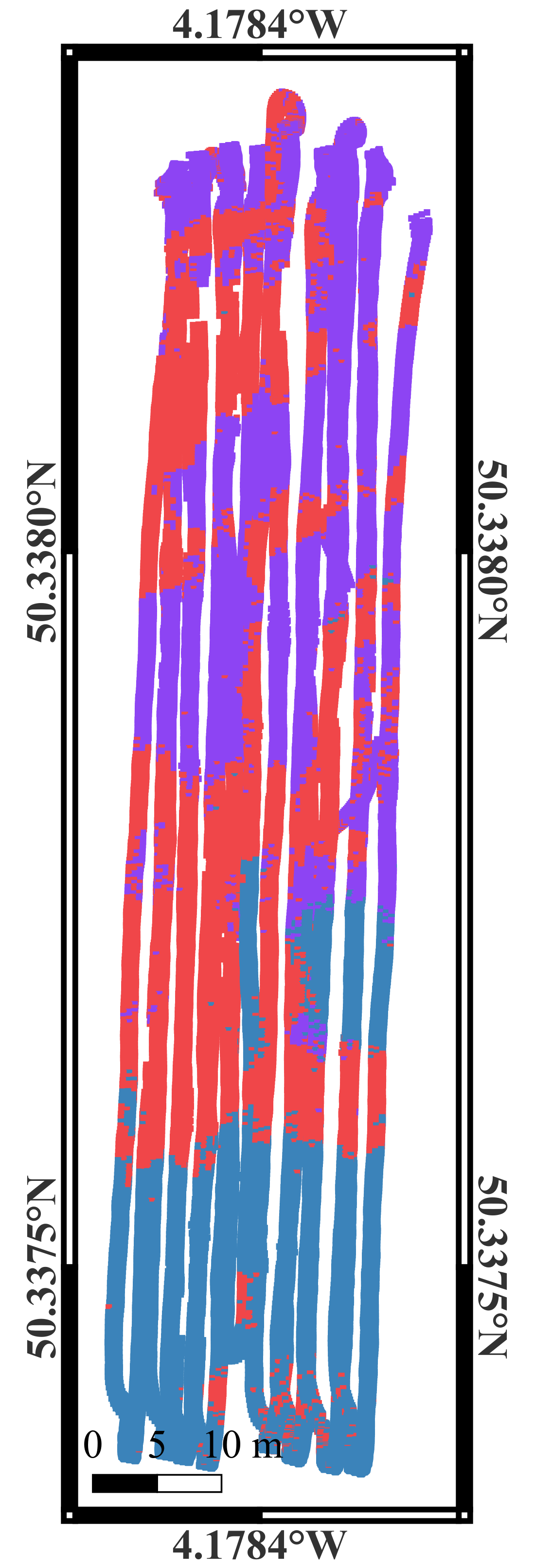}
    }
    \subfloat[Consistency map]{
        \includegraphics[width=0.3\linewidth]{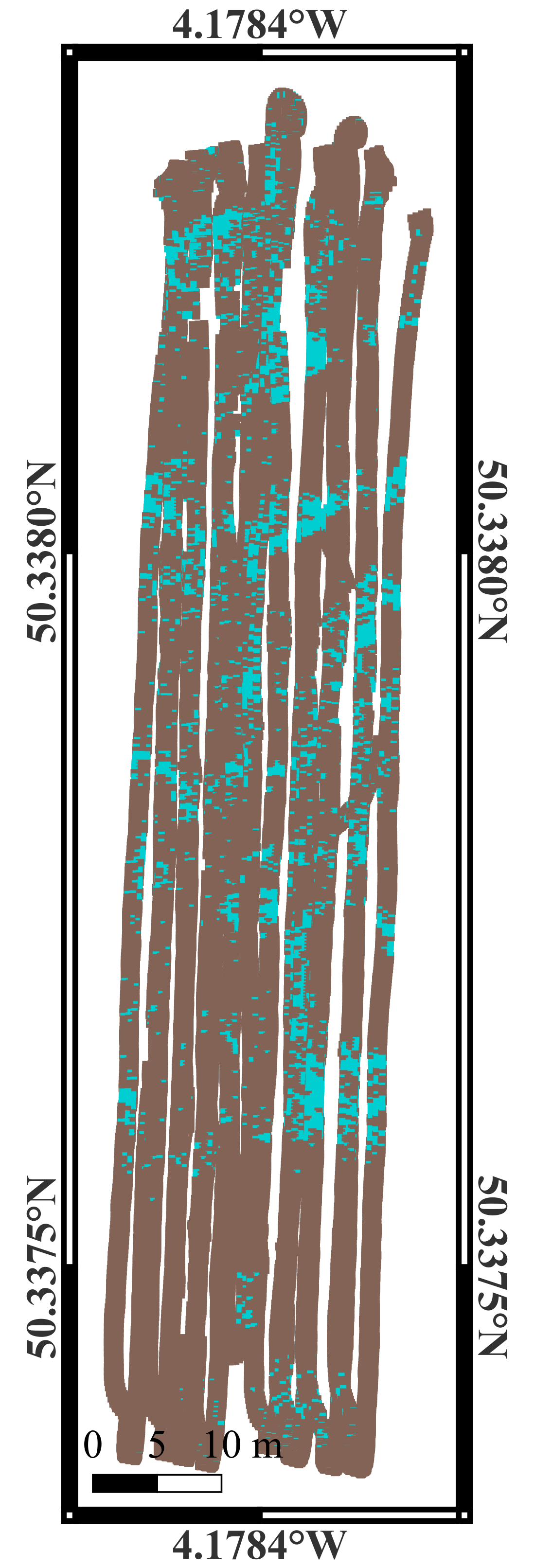}}
    \subfloat[Entropy map]{
        \includegraphics[width=0.3\linewidth]{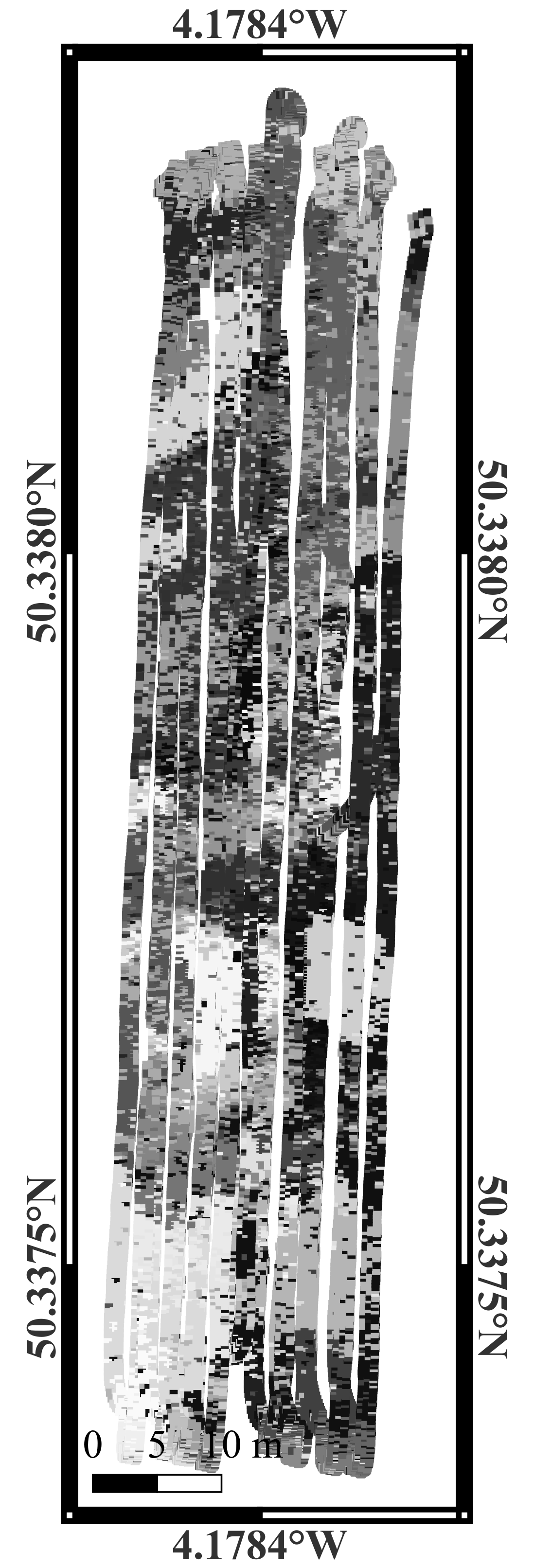}}
        
    \subfloat{\includegraphics[width=\linewidth]{photo/legend_classes.png}}
    \caption{Comparison of the ground truth and the inferred clustering results from the online clustering for the CB dataset. Panel (c) displays the entropy of each inferred cluster, which visualises the purity of the ground truth class distribution within that cluster.}
    \label{fig:cb_result}
\end{figure}

\begin{figure}
    \subfloat[Ground truth]{
        \includegraphics[width=0.48\linewidth]{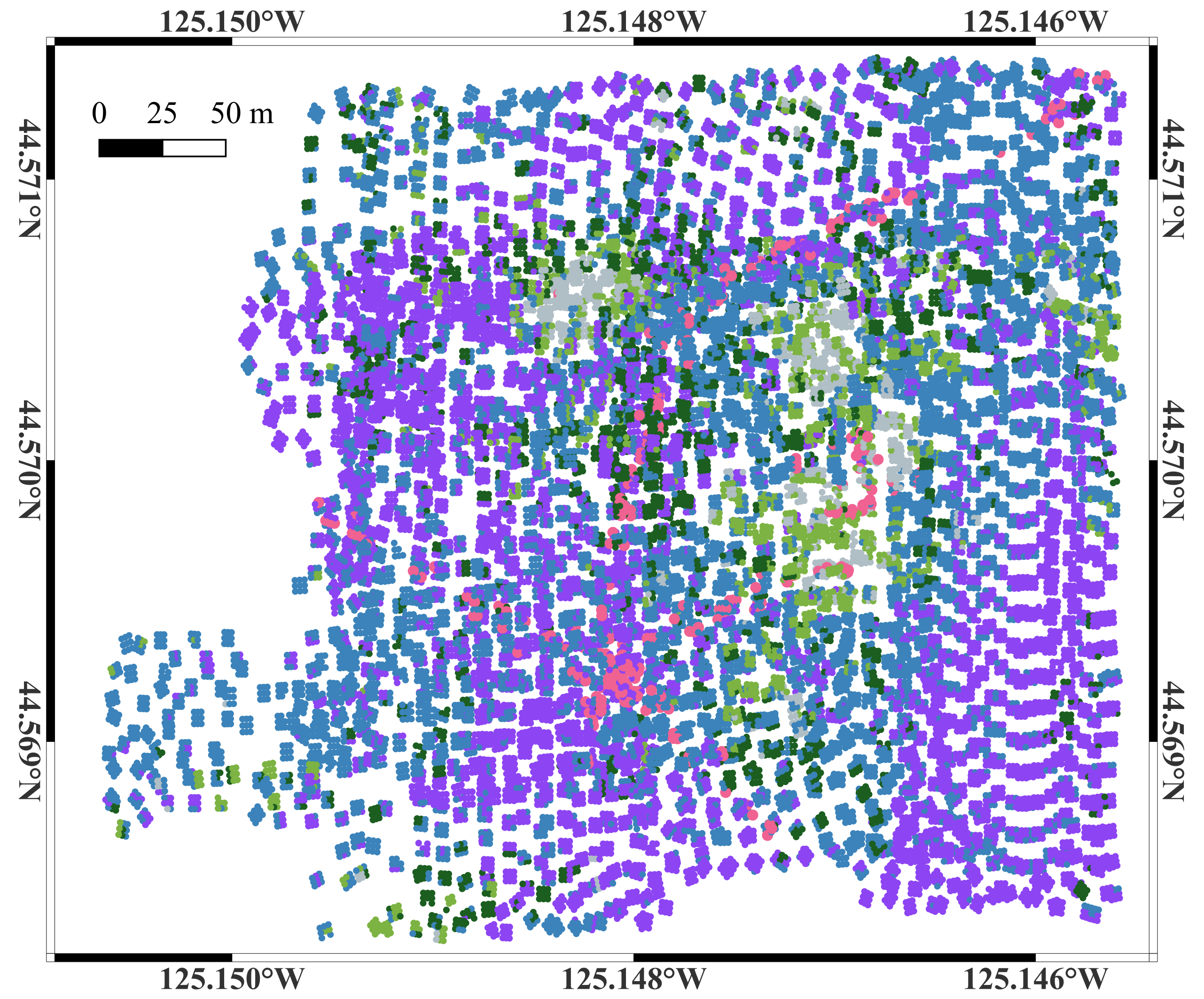}
    }
    \subfloat[Consistency map]{
        \includegraphics[width=0.48\linewidth]{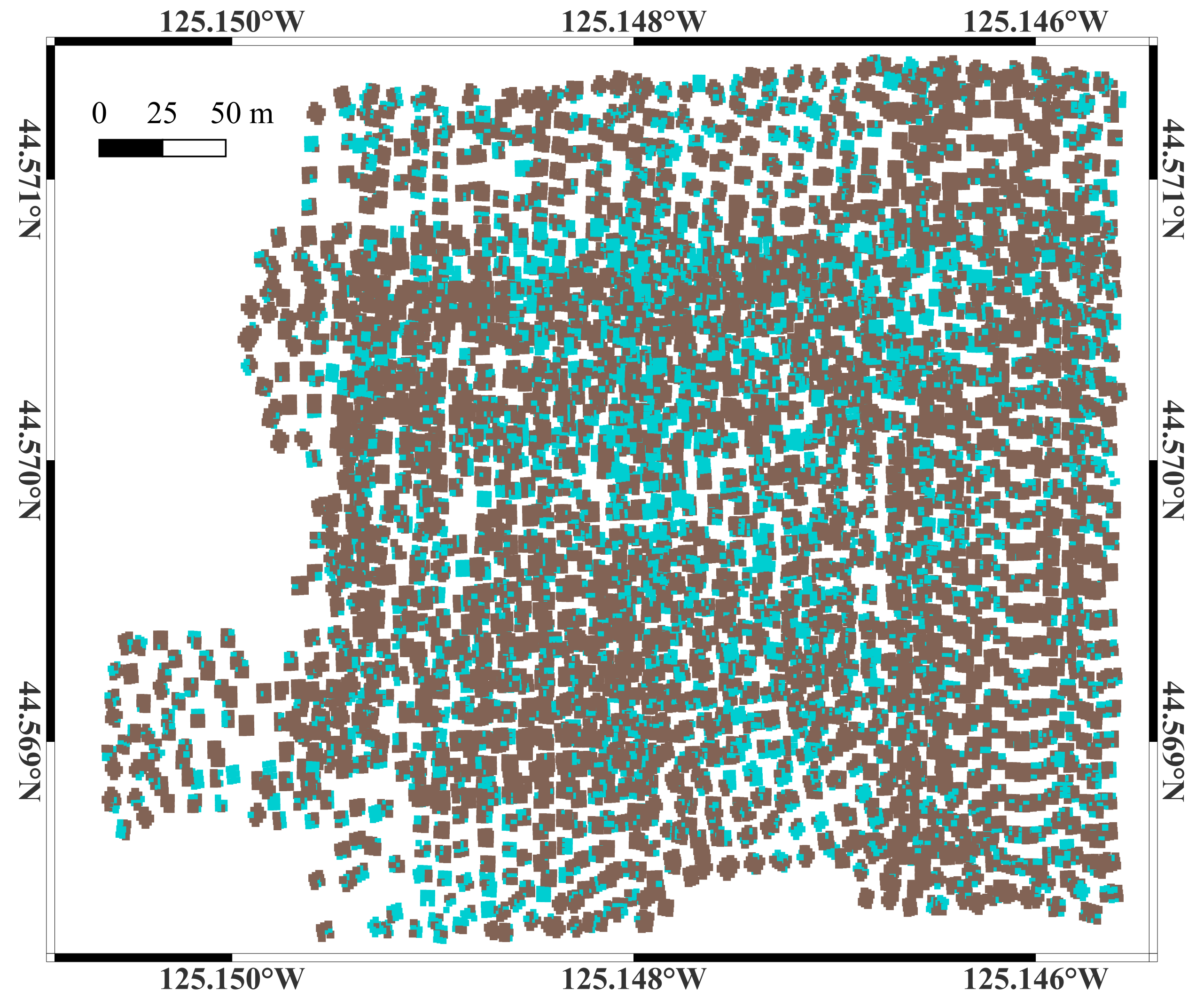}}
        
    \subfloat[Entropy map]{
        \includegraphics[width=0.48\linewidth]{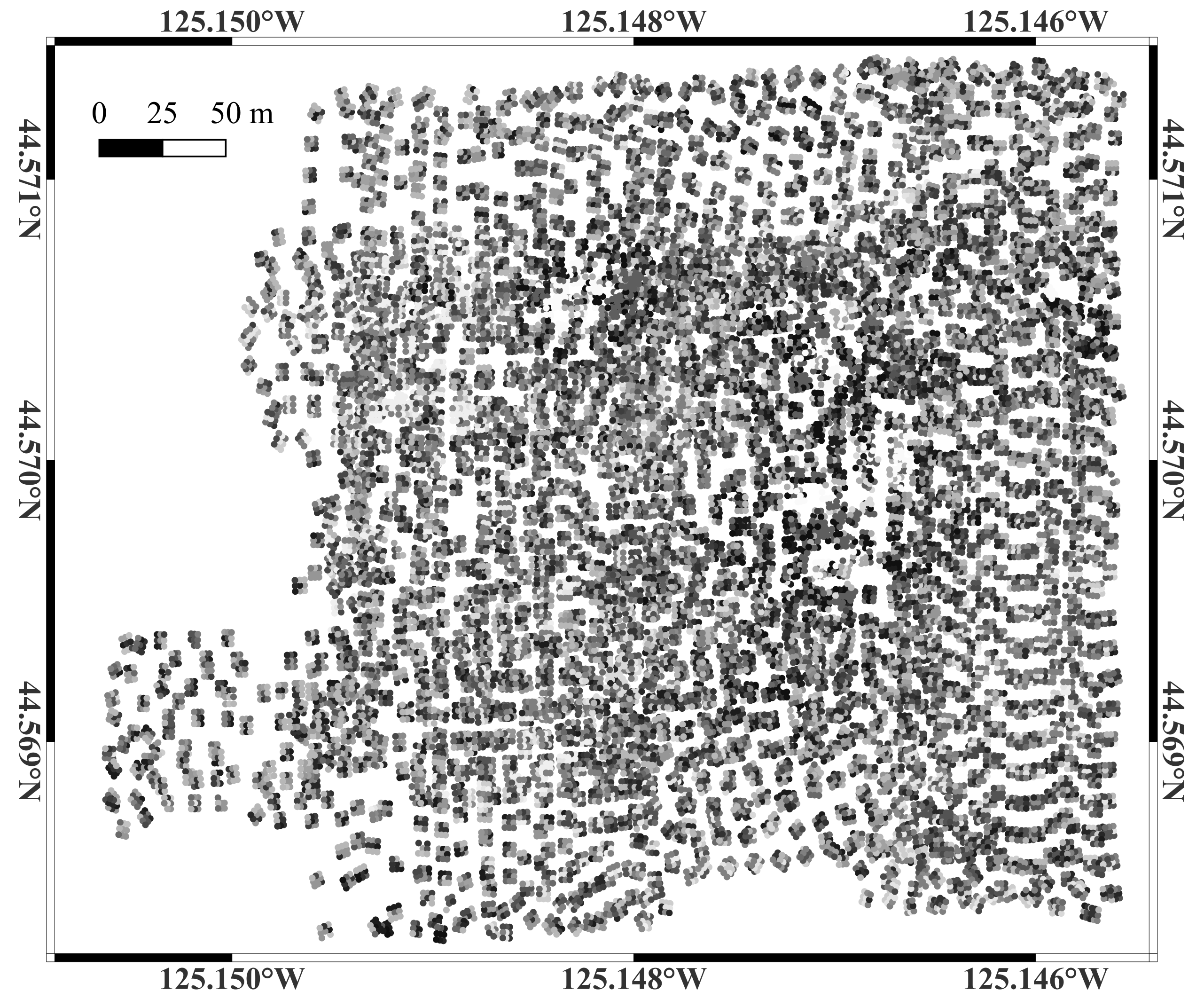}}
        
    \subfloat{\includegraphics[width=\linewidth]{photo/legend_classes.png}}
    \caption{Comparison of the ground truth and the inferred clustering results from the online clustering for the SH dataset. Panel (c) displays the entropy of each inferred cluster, which visualises the purity of the ground truth class distribution within that cluster.}
    \label{fig:sh_result}
\end{figure}

Fig.~\ref{fig:dm_result}, \ref{fig:cb_result}, and \ref{fig:sh_result} show, for DM, CB and SH dataset, respectively, the ground-truth classes (colour scheme as in Fig.~\ref{fig:representative_images}), consistency maps (b) highlighting disagreements between ground truth and \textit{OC + Density-based Sampling}, and entropy maps (c) indicating within-cluster class mixing extent (high entropy $\sim$ heterogeneous; low $\sim$
class-dominant). Most errors occur in habitat transition zones. In DM, Coral Mound Edge and Tail are frequently confused; entropy peaks around the Coral Mound Edge indicate residual ambiguity even when Coral Mound Top is majority-labelled. In CB, discrepancies are stronger, concentrated along Rock–Sediment and Rock–Fine Gravel boundaries, with high-entropy regions co-located with disagreements. In SH, errors concentrate in Shell Fragment, rarely recovered by majority voting in clustering results, reflecting weak separability and visual similarity to other habitats; by contrast, Cable, Rock, and Bacterial Mat are reliably inferred.

\section{Conclusion}
The online clustering framework is developed to enable real-time or near-real-time interpretation of seafloor images during long-term AUV deployments to facilitate downstream tasks relying on in-situ images, where data is continuously acquired and the volume of collected images increases over time. The framework integrates key components such as dynamic cluster merging and splitting, representative sample selection, and backbone models, each of which demonstrates clear advantages through comparative experiments on three field survey datasets. The framework achieves an average F1 score of $0.678 \pm 0.0147$ across three datasets with various observing trajectories, demonstrating clustering performance that is comparable to offline methods, which are time-consuming, and superior to other online clustering baselines. 
Therefore, it is well-suited for onboard implementation as a perception–recognition module to support downstream tasks during practical AUV deployments.

\section*{Acknowledgment}
The research was carried out under the Innovate UK OASIS project (10110715). Darwin Mounds data was gathered during NERC CLASS (NE/R015953/1) and Oceanids BioCam cruises of the RRS Discovery (NE/P020887/1 and NE/P020739/1). We thank Dr. Veerle Huvenne for assisting in generating human-verified labels for the Darwin Mounds dataset. The Southern Hydrate Ridge data was gathered during the Schmidt Ocean Institute's \#AdaptiveRobotics campaign (FK180731). The Cawsand Bay dataset was gathered using the University of Southampton's Smarty200 AUV (EPSRC EP/V035975/1). Cailei Liang is funded by the China Scholarship Council.

\bibliographystyle{IEEEtran}
\bibliography{IEEEabrv,Bibliography}

\begin{thebibliography}{10}
\providecommand{\url}[1]{#1}
\csname url@rmstyle\endcsname
\providecommand{\newblock}{\relax}
\providecommand{\bibinfo}[2]{#2}
\providecommand\BIBentrySTDinterwordspacing{\spaceskip=0pt\relax}
\providecommand\BIBentryALTinterwordstretchfactor{4}
\providecommand\BIBentryALTinterwordspacing{\spaceskip=\fontdimen2\font plus
\BIBentryALTinterwordstretchfactor\fontdimen3\font minus \fontdimen4\font\relax}
\providecommand\BIBforeignlanguage[2]{{%
\expandafter\ifx\csname l@#1\endcsname\relax
\typeout{** WARNING: IEEEtran.bst: No hyphenation pattern has been}%
\typeout{** loaded for the language `#1'. Using the pattern for}%
\typeout{** the default language instead.}%
\else
\language=\csname l@#1\endcsname
\fi
#2}}

\bibitem{williams}
S.~B. Williams, O.~R. Pizarro, M.~V. Jakuba, C.~R. Johnson, N.~S. Barrett, R.~C. Babcock, G.~A. Kendrick, P.~D. Steinberg, A.~J. Heyward, P.~J. Doherty, I.~Mahon, M.~Johnson-Roberson, D.~Steinberg, and A.~Friedman, ``Monitoring of benthic reference sites: Using an autonomous underwater vehicle,'' \emph{IEEE Robotics \& Automation Magazine}, vol.~19, no.~1, pp. 73--84, 2012.

\bibitem{nauert2023inspection}
F.~Nauert and P.~Kampmann, ``Inspection and maintenance of industrial infrastructure with autonomous underwater robots,'' \emph{Frontiers in Robotics and AI}, vol.~10, p. 1240276, 2023.

\bibitem{singh}
\BIBentryALTinterwordspacing
R.~M. Eustice, H.~Singh, J.~J. Leonard, and M.~R. Walter, ``Visually mapping the rms titanic: Conservative covariance estimates for slam information filters,'' \emph{The International Journal of Robotics Research}, vol.~25, no.~12, pp. 1223--1242, 2006. [Online]. Available: \url{https://doi.org/10.1177/0278364906072512}
\BIBentrySTDinterwordspacing

\bibitem{steinberg2015hierarchical}
D.~M. Steinberg, O.~Pizarro, and S.~B. Williams, ``Hierarchical bayesian models for unsupervised scene understanding,'' \emph{Computer Vision and Image Understanding}, vol. 131, pp. 128--144, 2015.

\bibitem{yamada2021geoclr}
T.~Yamada, A.~Pr{\"u}gel-Bennett, S.~B. Williams, O.~Pizarro, and B.~Thornton, ``Geoclr: Georeference contrastive learning for efficient seafloor image interpretation,'' \emph{arXiv preprint arXiv:2108.06421}, 2021.

\bibitem{bodenmann2017generation}
A.~Bodenmann, B.~Thornton, and T.~Ura, ``Generation of high-resolution three-dimensional reconstructions of the seafloor in color using a single camera and structured light,'' \emph{Journal of Field Robotics}, vol.~34, no.~5, pp. 833--851, 2017.

\bibitem{zelada}
\BIBentryALTinterwordspacing
A.~Zelada~Leon, V.~A. Huvenne, N.~M. Benoist, M.~Ferguson, B.~J. Bett, and R.~B. Wynn, ``Assessing the repeatability of automated seafloor classification algorithms, with application in marine protected area monitoring,'' \emph{Remote Sensing}, vol.~12, no.~10, 2020. [Online]. Available: \url{https://www.mdpi.com/2072-4292/12/10/1572}
\BIBentrySTDinterwordspacing

\bibitem{hwang2019auv}
J.~Hwang, N.~Bose, and S.~Fan, ``Auv adaptive sampling methods: A review,'' \emph{Applied Sciences}, vol.~9, no.~15, p. 3145, 2019.

\bibitem{murphy2013}
C.~Murphy, J.~M. Walls, T.~Schneider, R.~M. Eustice, M.~Stojanovic, and H.~Singh, ``Capture: A communications architecture for progressive transmission via underwater relays with eavesdropping,'' \emph{IEEE Journal of Oceanic Engineering}, vol.~39, no.~1, pp. 120--130, 2014.

\bibitem{oquab2023dinov2}
M.~Oquab, T.~Darcet, T.~Moutakanni, H.~Vo, M.~Szafraniec, V.~Khalidov, P.~Fernandez, D.~Haziza, F.~Massa, A.~El-Nouby, \emph{et~al.}, ``Dinov2: Learning robust visual features without supervision,'' \emph{arXiv preprint arXiv:2304.07193}, 2023.

\bibitem{catami}
\BIBentryALTinterwordspacing
F.~Althaus, N.~Hill, R.~Ferrari, L.~Edwards, R.~Przeslawski, C.~H.~L. Schönberg, R.~Stuart-Smith, N.~Barrett, G.~Edgar, J.~Colquhoun, M.~Tran, A.~Jordan, T.~Rees, and K.~Gowlett-Holmes, ``A standardised vocabulary for identifying benthic biota and substrata from underwater imagery: The catami classification scheme,'' \emph{PLOS ONE}, vol.~10, no.~10, pp. 1--18, 10 2015. [Online]. Available: \url{https://doi.org/10.1371/journal.pone.0141039}
\BIBentrySTDinterwordspacing

\bibitem{Langenkamper}
\BIBentryALTinterwordspacing
D.~Langenkämper, R.~van Kevelaer, A.~Purser, and T.~W. Nattkemper, ``Gear-induced concept drift in marine images and its effect on deep learning classification,'' \emph{Frontiers in Marine Science}, vol. Volume 7 - 2020, 2020. [Online]. Available: \url{https://www.frontiersin.org/journals/marine-science/articles/10.3389/fmars.2020.00506}
\BIBentrySTDinterwordspacing

\bibitem{yamada2022guiding}
T.~Yamada, M.~Massot-Campos, A.~Pr{\"u}gel-Bennett, O.~Pizarro, S.~B. Williams, and B.~Thornton, ``Guiding labelling effort for efficient learning with georeferenced images,'' \emph{IEEE Transactions on Pattern Analysis and Machine Intelligence}, vol.~45, no.~1, pp. 593--607, 2022.

\bibitem{ikotun2023k}
A.~M. Ikotun, A.~E. Ezugwu, L.~Abualigah, B.~Abuhaija, and J.~Heming, ``K-means clustering algorithms: A comprehensive review, variants analysis, and advances in the era of big data,'' \emph{Information Sciences}, vol. 622, pp. 178--210, 2023.

\bibitem{bhattacharjee2021survey}
P.~Bhattacharjee and P.~Mitra, ``A survey of density based clustering algorithms,'' \emph{Frontiers of Computer Science}, vol.~15, pp. 1--27, 2021.

\bibitem{wu2022modeling}
B.~Wu, S.~Sakti, J.~Zhang, and S.~Nakamura, ``Modeling unsupervised empirical adaptation by dpgmm and dpgmm-rnn hybrid model to extract perceptual features for low-resource asr,'' \emph{IEEE/ACM Transactions on Audio, Speech, and Language Processing}, vol.~30, pp. 901--916, 2022.

\bibitem{dino}
M.~Caron, H.~Touvron, I.~Misra, H.~J\'{e}gou, J.~Mairal, P.~Bojanowski, and A.~Joulin, ``Emerging properties in self-supervised vision transformers,'' \emph{Proceedings of the IEEE/CVF International Conference on Computer Vision (ICCV)}, pp. 9650--9660, 2021.

\bibitem{yamada2021learning}
T.~Yamada, A.~Pr{\"u}gel-Bennett, and B.~Thornton, ``Learning features from georeferenced seafloor imagery with location guided autoencoders,'' \emph{Journal of Field Robotics}, vol.~38, no.~1, pp. 52--67, 2021.

\bibitem{kudithipudi2022biological}
D.~Kudithipudi, M.~Aguilar-Simon, J.~Babb, M.~Bazhenov, D.~Blackiston, J.~Bongard, A.~P. Brna, S.~Chakravarthi~Raja, N.~Cheney, J.~Clune, \emph{et~al.}, ``Biological underpinnings for lifelong learning machines,'' \emph{Nature Machine Intelligence}, vol.~4, no.~3, pp. 196--210, 2022.

\bibitem{sun2021and}
G.~Sun, Y.~Cong, J.~Dong, Y.~Liu, Z.~Ding, and H.~Yu, ``What and how: generalized lifelong spectral clustering via dual memory,'' \emph{IEEE transactions on pattern analysis and machine intelligence}, vol.~44, no.~7, pp. 3895--3908, 2021.

\bibitem{wu2019large}
Y.~Wu, Y.~Chen, L.~Wang, Y.~Ye, Z.~Liu, Y.~Guo, and Y.~Fu, ``Large scale incremental learning,'' in \emph{Proceedings of the IEEE/CVF conference on computer vision and pattern recognition}, 2019, pp. 374--382.

\bibitem{rao2019continual}
D.~Rao, F.~Visin, A.~Rusu, R.~Pascanu, Y.~W. Teh, and R.~Hadsell, ``Continual unsupervised representation learning,'' \emph{Advances in neural information processing systems}, vol.~32, 2019.

\bibitem{rolnick2019experience}
D.~Rolnick, A.~Ahuja, J.~Schwarz, T.~Lillicrap, and G.~Wayne, ``Experience replay for continual learning,'' in \emph{Advances in Neural Information Processing Systems (NeurIPS)}, 2019, pp. 350--360.

\bibitem{yang2022lifelong}
Z.~Yang, J.~Zheng, and Z.~Ge, ``Lifelong bayesian learning machines for streaming industrial big data,'' \emph{IEEE Transactions on Systems, Man, and Cybernetics: Systems}, vol.~53, no.~3, pp. 1554--1565, 2022.

\bibitem{hu2013incremental}
W.~Hu, X.~Li, G.~Tian, S.~Maybank, and Z.~Zhang, ``An incremental dpmm-based method for trajectory clustering, modeling, and retrieval,'' \emph{IEEE transactions on pattern analysis and machine intelligence}, vol.~35, no.~5, pp. 1051--1065, 2013.

\bibitem{chaudhry2019tiny}
A.~Chaudhry, M.~Rohrbach, M.~Elhoseiny, T.~Ajanthan, P.~H. Torr, and P.~K. Dokania, ``Tiny episodic memories in continual learning,'' in \emph{Proceedings of the International Conference on Machine Learning (ICML)}, 2019, pp. 1954--1963.

\bibitem{kim2019incremental}
D.~Kim, J.~Bae, Y.~Jo, and J.~Choi, ``Incremental learning with maximum entropy regularization: Rethinking forgetting and intransigence,'' \emph{arXiv preprint arXiv:1902.00829}, 2019.

\bibitem{gou2021knowledge}
J.~Gou, B.~Yu, S.~J. Maybank, and D.~Tao, ``Knowledge distillation: A survey,'' \emph{International Journal of Computer Vision}, vol. 129, no.~6, pp. 1789--1819, 2021.

\bibitem{hinton2015distilling}
G.~Hinton, O.~Vinyals, and J.~Dean, ``Distilling the knowledge in a neural network,'' \emph{arXiv preprint arXiv:1503.02531}, 2015.

\bibitem{asif2019ensemble}
U.~Asif, J.~Tang, and S.~Harrer, ``Ensemble knowledge distillation for learning improved and efficient networks,'' \emph{arXiv preprint arXiv:1909.08097}, 2019.

\bibitem{kim2021feature}
J.~Kim, M.~Hyun, I.~Chung, and N.~Kwak, ``Feature fusion for online mutual knowledge distillation,'' in \emph{2020 25th International Conference on Pattern Recognition (ICPR)}.\hskip 1em plus 0.5em minus 0.4em\relax IEEE, 2021, pp. 4619--4625.

\bibitem{tasar2019incremental}
O.~Tasar, Y.~Tarabalka, and P.~Alliez, ``Incremental learning for semantic segmentation of large-scale remote sensing data,'' \emph{IEEE Journal of Selected Topics in Applied Earth Observations and Remote Sensing}, vol.~12, no.~9, pp. 3524--3537, 2019.

\bibitem{lughofer2015autonomous}
E.~Lughofer and M.~Sayed-Mouchaweh, ``Autonomous data stream clustering implementing split-and-merge concepts--towards a plug-and-play approach,'' \emph{Information Sciences}, vol. 304, pp. 54--79, 2015.

\bibitem{dinari2022sampling}
O.~Dinari and O.~Freifeld, ``Sampling in dirichlet process mixture models for clustering streaming data,'' in \emph{International Conference on Artificial Intelligence and Statistics}.\hskip 1em plus 0.5em minus 0.4em\relax PMLR, 2022, pp. 818--835.

\bibitem{ronen2022deepdpm}
M.~Ronen, S.~E. Finder, and O.~Freifeld, ``Deepdpm: Deep clustering with an unknown number of clusters,'' in \emph{Proceedings of the IEEE/CVF Conference on Computer Vision and Pattern Recognition}, 2022, pp. 9861--9870.

\bibitem{ding2002cluster}
C.~Ding and X.~He, ``Cluster merging and splitting in hierarchical clustering algorithms,'' in \emph{2002 IEEE International Conference on Data Mining, 2002. Proceedings.}\hskip 1em plus 0.5em minus 0.4em\relax IEEE, 2002, pp. 139--146.

\bibitem{wagenmakers2004aic}
E.-J. Wagenmakers and S.~Farrell, ``Aic model selection using akaike weights,'' \emph{Psychonomic bulletin \& review}, vol.~11, pp. 192--196, 2004.

\bibitem{creswell2018generative}
A.~Creswell, T.~White, V.~Dumoulin, K.~Arulkumaran, B.~Sengupta, and A.~A. Bharath, ``Generative adversarial networks: An overview,'' \emph{IEEE signal processing magazine}, vol.~35, no.~1, pp. 53--65, 2018.

\bibitem{yu2023dataset}
R.~Yu, S.~Liu, and X.~Wang, ``Dataset distillation: A comprehensive review,'' \emph{IEEE Transactions on Pattern Analysis and Machine Intelligence}, 2023.

\bibitem{lin2022anchor}
H.~Lin, S.~Feng, X.~Li, W.~Li, and Y.~Ye, ``Anchor assisted experience replay for online class-incremental learning,'' \emph{IEEE Transactions on Circuits and Systems for Video Technology}, 2022.

\bibitem{sachdeva2023data}
N.~Sachdeva and J.~McAuley, ``Data distillation: A survey,'' \emph{arXiv preprint arXiv:2301.04272}, 2023.

\bibitem{lei2023comprehensive}
S.~Lei and D.~Tao, ``A comprehensive survey of dataset distillation,'' \emph{IEEE Transactions on Pattern Analysis and Machine Intelligence}, 2023.

\bibitem{grafstrom2012spatially}
A.~Grafstr{\"o}m, N.~L. Lundstr{\"o}m, and L.~Schelin, ``Spatially balanced sampling through the pivotal method,'' \emph{Biometrics}, vol.~68, no.~2, pp. 514--520, 2012.

\bibitem{dubin1992spatial}
R.~A. Dubin, ``Spatial autocorrelation and neighborhood quality,'' \emph{Regional science and urban economics}, vol.~22, no.~3, pp. 433--452, 1992.

\bibitem{blatchford2021determining}
M.~L. Blatchford, C.~M. Mannaerts, and Y.~Zeng, ``Determining representative sample size for validation of continuous, large continental remote sensing data,'' \emph{International Journal of Applied Earth Observation and Geoinformation}, vol.~94, p. 102235, 2021.

\bibitem{daszykowski2002representative}
M.~Daszykowski, B.~Walczak, and D.~Massart, ``Representative subset selection,'' \emph{Analytica chimica acta}, vol. 468, no.~1, pp. 91--103, 2002.

\bibitem{broderick2012beta}
T.~Broderick, M.~I. Jordan, and J.~Pitman, ``Beta processes, stick-breaking and power laws,'' 2012.

\bibitem{lughofer2012dynamic}
E.~Lughofer, ``A dynamic split-and-merge approach for evolving cluster models,'' \emph{Evolving systems}, vol.~3, no.~3, pp. 135--151, 2012.

\bibitem{caron2020unsupervised}
M.~Caron, I.~Misra, J.~Mairal, P.~Goyal, P.~Bojanowski, and A.~Joulin, ``Unsupervised learning of visual features by contrasting cluster assignments,'' \emph{Advances in neural information processing systems}, vol.~33, pp. 9912--9924, 2020.

\bibitem{liang2025self}
C.~Liang, J.~Cappelletto, M.~Massot-Campos, A.~Bodenmann, V.~A. Huvenne, C.~Wardell, B.~J. Bett, D.~Newborough, and B.~Thornton, ``Self-supervised learning with multimodal remote sensed maps for seafloor visual class inference,'' \emph{The International Journal of Robotics Research}, p. 02783649251343640, 2025.

\bibitem{Liang2025LRSSL}
C.~Liang, A.~Bodenmann, E.~J. Curtis, S.~Simmons, K.~Nagano, S.~Brown, A.~Riese, and B.~Thornton, ``Investigating location-regularised self-supervised feature learning for seafloor visual imagery,'' 2025, submitted to IEEE Journal of Oceanic Engineering.

\end{thebibliography}



\end{document}